\documentclass[fontB]{framexpaper}

\usepackage{multirow,pifont,xspace}
\usepackage[capitalise]{cleveref}
\definecolor{lightblue}{RGB}{232,244,255}
\definecolor{paleRed}{RGB}{255,204,204}
\definecolor{palePurple}{RGB}{210,200,250}
\newcommand{\cmark}{\ding{51}}
\newcommand{\xmark}{\ding{55}}

\makeatletter
\patchcmd{\pgf@sys@pdf@install@mask}
  {\expandafter\let\csname pgfsmask@#1\endcsname\expandafter\csname pgfsmaskxform@#1\endcsname}
  {\csletcs{pgfsmask@#1}{pgfsmaskxform@#1}}{}{}
\DeclareRobustCommand{\onedot}{\futurelet\@let@token\@onedot}
\def\@onedot{\ifx\@let@token.\else.\null\fi\xspace}
\makeatother
\def\eg{\emph{e.g}\onedot}

\usepackage[section]{placeins}
\usepackage{wrapfig}

\NewDocumentEnvironment{fffigure}{O{htbp}}{%
  \begin{figure}[#1]\centering
  \captionsetup{width=.94\linewidth,position=bottom}%
}{\end{figure}}
\NewDocumentEnvironment{fftable}{O{htbp}}{%
  \begin{table}[#1]\centering
  \rmfamily\small
  \setlength{\tabcolsep}{6pt}%
  \arrayrulecolor{black}%
  \captionsetup{width=.94\linewidth,position=top,
    justification=centering,singlelinecheck=false}%
}{\end{table}}
\NewDocumentEnvironment{fftabular}{m +b}{%
  \begin{tabular}{#1}#2\end{tabular}%
}{}
\newcommand{\fftablenote}[1]{%
  \par\vspace{4pt}%
  {\rmfamily\small\centering #1\par}}
\newcommand{\ffresultrow}{\rowcolor{FXPale}}

\title{\textcolor{FXForest}{FrameForge:} Harnessing Priors of Video Foundation Models for Unified Image Generation}
\author{Yijing Lin\textsuperscript{1}\quad Zhendong Mao\textsuperscript{1,\fxemailmark}\quad Mengqi Huang\textsuperscript{1,2}\\[2pt]
Yuxin Wang\textsuperscript{2}\quad Yongdong Zhang\textsuperscript{1}}
\renewcommand\fxaffiliations{\textsuperscript{1}University of Science and Technology of China\quad\textsuperscript{2}FrameX.AI}
\renewcommand\fxauthornote{%
  \fxemaillink{zdmao@ustc.edu.cn}\enspace Corresponding author: Zhendong Mao\\[2pt]
  \href{mailto:lyijing@mail.ustc.edu.cn}{lyijing@mail.ustc.edu.cn};
  \href{mailto:zdmao@ustc.edu.cn}{zdmao@ustc.edu.cn};
  \href{mailto:huangmq@ustc.edu.cn}{huangmq@ustc.edu.cn}\\[2pt]
  \href{mailto:wangyuxin@frame-x.ai}{wangyuxin@frame-x.ai};
  \href{mailto:zhyd73@ustc.edu.cn}{zhyd73@ustc.edu.cn}}
\renewcommand\fxprojectpageurl{https://lyne1.github.io/FrameForge-page/}
\renewcommand\fxprojectpagedisplay{lyne1.github.io/FrameForge-page/}
\renewcommand\fxprojectlinks{%
  \par\noindent
  \begin{tikzpicture}[x=1pt,y=1pt,baseline=(fxlabel.base)]
    \path[use as bounding box] (0pt,-5.2pt) rectangle (\linewidth,8.8pt);
    \node[anchor=center,inner sep=0pt] at (7pt,1.75pt)
      {\includegraphics[width=14pt,height=12.6pt,keepaspectratio]{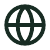}};
    \node[anchor=base west,inner sep=0pt,font=\sffamily\fontsize{8.8}{10.8}\selectfont\bfseries,
      text=FXForest] (fxlabel) at (21pt,0pt) {Project Page:};
    \node[anchor=base west,inner sep=0pt,font=\sffamily\fontsize{8.8}{10.8}\selectfont,
      text=FXURL] at ([xshift=2pt]fxlabel.base east)
      {\href{\fxprojectpageurl}{\fxprojectpagedisplay}};
  \end{tikzpicture}\par\vspace{4.5pt}}

\hypersetup{
  pdftitle={FrameForge: Harnessing Priors of Video Foundation Models for Unified Image Generation},
  pdfauthor={Yijing Lin; Zhendong Mao; Mengqi Huang; Yuxin Wang; Yongdong Zhang},
  pdfsubject={Unified conditional image generation}}

\begin{document}
\maketitle
\begin{fxabstract}
Unified conditional image generation requires a single model to preserve different aspects of a visual input while producing a target image that follows textual guidance. Pretrained video models provide a promising foundation by encoding visual transformations across time, yet existing approaches typically assign the condition and target to fixed temporal positions. We observe that condition-target relationships vary substantially across generation tasks, while video models naturally encode visual transformations at different temporal scales. This motivates us to model conditional generation through input-adaptive virtual temporal relations. We propose \textbf{FrameForge}, which captures two complementary aspects of such relations. \textbf{Mixture of Temporal Experts (MoTE)} models \emph{how far} the target lies from the condition by dynamically selecting temporal RoPE experts with different virtual intervals. \textbf{Chain of Temporal Reasoning (CoTR)} models \emph{how} the condition is transformed into the target through a latent intermediate state, with a reasoning-bottleneck attention mask encouraging information propagation through this pathway. Experiments on DreamBench, COCO2017, and GEdit-Bench show that FrameForge achieves superior performance in subject-driven generation,  canny-to-image generation, and image editing among the compared methods.

\end{fxabstract}

\section{Introduction}\label{sec:introduction}
With recent advances in generative modeling~\cite{ho2020denoising,huang2023towards,xu2024prompt,liu2025one,liu2026lacon,zhao2026condfoodgen,chen2026detail++}, foundation models such as Stable Diffusion~\cite{esser2021taming,podell2023sdxl,esser2024scalingrectifiedflowtransformers} and FLUX~\cite{flux2024} offer strong text-to-image generation capabilities. However, real-world generation often involves diverse inputs (\eg, text, reference images, and structural maps) that impose constraints at different semantic and spatial levels. Modeling these heterogeneous condition-target relations within a unified framework remains challenging.

\begin{wrapfigure}{R}{.48\textwidth}
  \captionsetup{width=.94\linewidth,position=bottom}
      \centering 
          \includegraphics[width=\linewidth,pagebox=cropbox]{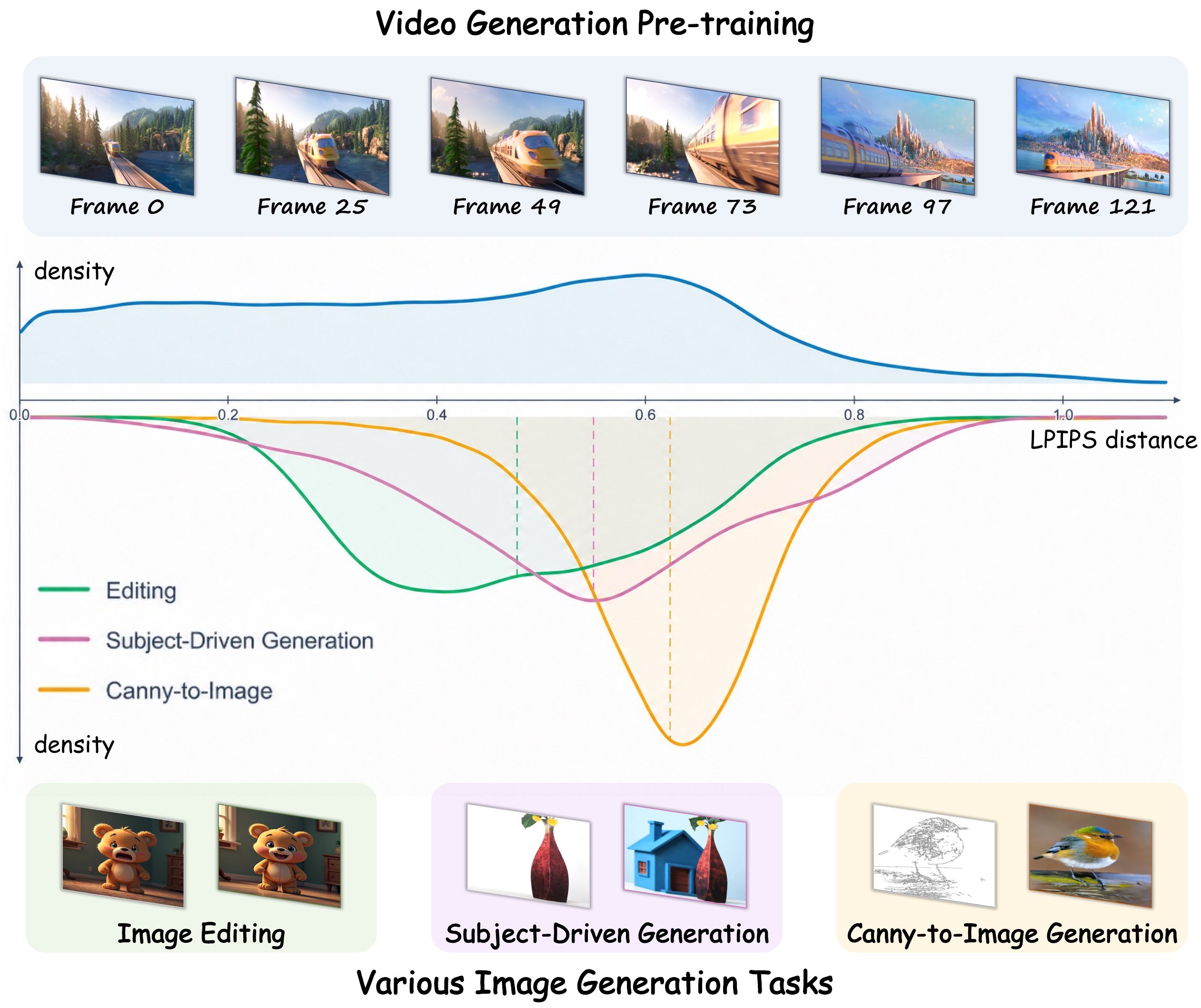}
    \caption{Motivation of FrameForge. 
    Top: The LPIPS distances between frame pairs sampled from natural videos at different temporal intervals. Bottom: The condition-to-target LPIPS distances for image editing, subject-driven generation, and canny-to-image generation, where the dashed lines denote the median distance of each task.}
    \label{fig:motivation}
\end{wrapfigure}

Existing approaches to heterogeneous conditional image generation are predominantly image-centric. Some train unified models on large-scale multi-task datasets~\cite{qin2023unicontrol,bai2024sequential,sun2024x,chen2024unireal,deng2025bagel,xiao2025omnigen,wu2025omnigen2}, while others extend pretrained generators with control branches or lightweight adapters~\cite{zhang2023adding,ye2023ip,yu2024anyedit,tan2024ominicontrol,zhang2025easycontrol}. These approaches typically inject conditions as auxiliary spatial or semantic signals without explicitly modeling how strongly the condition and target differ or how one should be transformed into the other. This can make it difficult for a shared model to accommodate substantially different generation behaviors.

Pretrained video generation models provide a natural alternative by learning visual transformations across time. Recent studies exploit these priors by arranging the condition and target as different frames~\cite{wu2025chronoedit,li2025video4editviewingimageediting,lin2025realgeneral,cao2026dimension}. However, existing approaches typically assign the condition and target to fixed temporal positions, resulting in an invariant temporal positional relation across different tasks and samples. Such a fixed offset overlooks the variation in condition-target relations and may limit a single shared adapter.

Our key observation is that heterogeneous condition-target relations can be interpreted through a \emph{virtual temporal interval}. In natural videos, frames separated by different temporal intervals exhibit different degrees of visual variation. A similar phenomenon arises in conditional image generation: image editing typically requires relatively small changes while preserving most source content, whereas subject-driven and canny-to-image generation often involve larger variations in appearance, structure, or scene composition. As shown in \cref{fig:motivation}, these tasks exhibit clearly different condition-target LPIPS distributions, while their variation ranges substantially overlap with those observed in natural video transitions. This suggests that different conditional generation tasks may correspond to different regions of the temporal transformation prior learned by a video model. The broad distributions within individual tasks further suggest that a task label alone cannot specify the required transformation magnitude. An input-dependent relation can account for both differences across tasks and variations among samples within the same task. Therefore, instead of assigning all condition-target pairs a fixed temporal offset, we model their relation with an input-adaptive virtual temporal interval.

This perspective reveals two complementary aspects of conditional generation: \emph{how far} the target lies from the condition under the learned temporal prior, and \emph{how} the condition should be transformed into the target. An adaptive interval addresses the former but does not uniquely determine the transformation. Modeling the latter requires a latent mechanism that learns condition-to-target transformations without explicit intermediate-frame supervision.

To this end, we propose \textbf{FrameForge}, a unified conditional image generation framework built upon a pretrained video foundation model. To model \emph{how far} the target lies from the condition, \textbf{Mixture of Temporal Experts (MoTE)} selects among temporal RoPE experts with different virtual intervals using the joint textual and visual context. To model \emph{how} the condition evolves into the target, \textbf{Chain of Temporal Reasoning (CoTR)} inserts a noise-initialized latent to form a condition-reasoning-target chain. The latent is progressively contextualized through attention with textual and visual tokens. During an intermediate training stage, a \textbf{reasoning-bottleneck attention mask} prevents target tokens from directly accessing text or condition tokens, encouraging information propagation through the reasoning latent. The MoTE router and a single shared LoRA are jointly optimized across all tasks, while the pretrained video backbone remains frozen. This design enables heterogeneous conditional generation without task-specific adapters.

Our main contributions are summarized as follows:

\textbf{Conceptual contribution.} We formulate heterogeneous conditional image generation from the perspective of input-adaptive virtual temporal relations in pretrained video models. Our analysis shows that different generation tasks exhibit distinct condition-target transformation magnitudes, motivating adaptive temporal relation modeling.

\textbf{Technical contribution.} We propose FrameForge, a unified adaptation framework that jointly models two complementary aspects of conditional generation. MoTE adaptively determines \emph{how far} the target lies from the condition through input-dependent temporal positional priors, while CoTR provides a latent pathway to model \emph{how} the condition is transformed into the target without explicit intermediate supervision. The entire framework is jointly trained across tasks with a single shared LoRA.

\textbf{Experimental contribution.} We conduct extensive experiments on subject-driven generation, canny-to-image generation, and instruction-guided image editing. Results on DreamBench, COCO2017, and GEdit demonstrate that a single unified FrameForge model achieves superior performance across all three tasks compared with task-specific and unified baselines. Comprehensive ablations further validate the effectiveness of the proposed MoTE and CoTR modules.
\section{Related Work}
\subsection{Unified Image Generation}
Conditional image generation includes structure-guided synthesis~\cite{zhang2023adding,mou2024t2i}, subject-driven generation~\cite{ruiz2023dreambooth,ye2023ip}, and instruction-guided editing~\cite{brooks2023instructpix2pix,zhang2023magicbrush}. Unified models use shared representations and parameters to preserve spatial layout, subject identity, or content outside the requested edit.

One line of research develops native unified models through large-scale multi-task training. PixWizard~\cite{lin2024pixwizard} adapts a flow-based diffusion transformer with textual and visual conditions. OmniGen~\cite{xiao2025omnigen} represents text, conditions, and targets as a multimodal sequence, while UniReal~\cite{chen2024unireal} treats input and output images as discontinuous video frames. OmniGen2~\cite{wu2025omnigen2} connects a vision-language model to a diffusion transformer through decoupled text and image pathways. BAGEL~\cite{deng2025bagel} extends unification to visual understanding through interleaved multimodal pretraining, and Lance~\cite{fu2026lanceunifiedmultimodalmodeling} combines shared context modeling, separate capability pathways, and staged image/video training.

Another line adapts pretrained image generators. OminiControl~\cite{tan2024ominicontrol} integrates condition-image tokens into a DiT's multimodal sequence and reuses its attention parameters for spatially aligned and non-aligned controls. ACE++~\cite{mao2025ace++} organizes references, masks, and target latents into unified context units. EasyControl~\cite{zhang2025easycontrol} uses isolated condition branches for plug-and-play integration and zero-shot composition, with position-aware training and KV caching for flexible resolution and efficient inference. Concurrent work, TAG-MoE~\cite{xu2026tagmoetaskawaregatingunified}, alleviates inter-task interference through sparse FFN experts and routing aligned with hierarchical task semantics. Unlike TAG-MoE, which routes image tokens to semantically specialized FFN experts, our MoTE retains shared feature-processing parameters and selects among temporal RoPE experts to model an input-dependent virtual temporal interval.

\subsection{Video Foundation Model Adaptation} 
Video diffusion models learn appearance and inter-frame variation through motion modules, as in AnimateDiff~\cite{guo2023animatediff}, or video diffusion transformers, as in Latte~\cite{ma2024latte}, SANA-Video~\cite{chen2025sana}, CogVideoX~\cite{yang2024cogvideox}, Wan~\cite{wan2025}, and HunyuanVideo~\cite{kong2024hunyuanvideo,hunyuanvideo2025}. Adapting these priors to image tasks requires relating the condition and target to the pretrained temporal structure.
 
RealGeneral~\cite{lin2025realgeneral} formulates conditional generation as two-frame prediction with unified conditional embeddings, decoupled modulation, and attention masking. DRA-Ctrl~\cite{cao2026dimension} uses mixup-based shot transitions to bridge continuous video dynamics and discontinuous image pairs. iMontage~\cite{fu2025iMontage} supports variable-length many-to-many generation by treating images as pseudo-frames and using Marginal RoPE to distinguish input and output sets from natural temporal ordering. Video4Edit~\cite{li2025video4editviewingimageediting} transfers intermediate evolution trajectories from a video-pretrained teacher to a student editor. ChronoEdit~\cite{wu2025chronoedit} targets physically consistent editing by jointly denoising the target and temporal reasoning tokens, which are discarded during later denoising steps.

Beyond task-specific adaptation, Wiedemer \emph{et al.}~\cite{wiedemer2025video} report emergent zero-shot abilities. Moreover, Acuaviva \emph{et al.}~\cite{acuaviva2025rethinking} examine these capabilities under controlled adaptation, providing evidence that video pretraining offers transferable priors for structured visual transformations.

Existing studies demonstrate the potential of video models as general foundations for image-level generation and visual reasoning. However, they are either tailored to individual tasks or fail to explicitly mitigate inter-task interference within a shared parameter-efficient adapter. Our framework addresses this issue by adapting the virtual temporal relation to each input and modeling the condition-to-target transformation.
\section{Method}
FrameForge adapts a pretrained video diffusion model to conditional image generation through two complementary operations: selecting an input-dependent temporal positional relation and learning an intermediate representation for condition-to-target prediction. We first introduce the preliminaries and framework overview in \cref{sec:preliminaries,sec:overview}, and then describe MoTE and CoTR in \cref{sec:mote,sec:cotr}. The training stages are detailed in \cref{sec:lora}.

\begin{fffigure}[htbp]
  \centering
  \includegraphics[width=\linewidth,pagebox=cropbox]{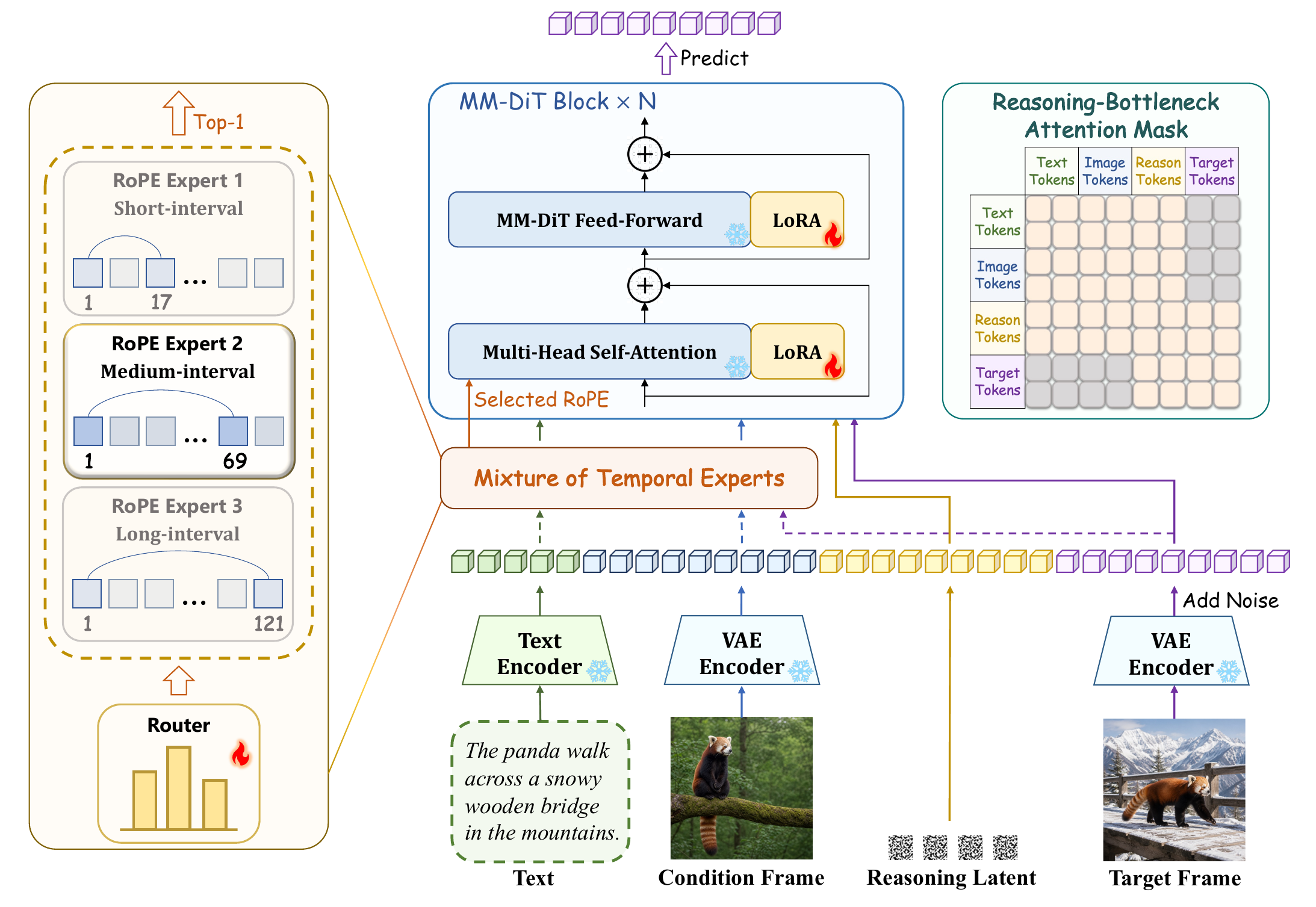}
  \caption{Overview of FrameForge. Heterogeneous conditional image generation is formulated as target-frame prediction with a pretrained video diffusion model. The MM-DiT backbone processes text, condition, reasoning, and noisy target tokens to predict the target velocity. Mixture of Temporal Experts (MoTE) selects a temporal RoPE configuration at each denoising step, adapting the virtual interval between the condition and target. Chain of Temporal Reasoning (CoTR) adds a noise-initialized intermediate latent whose hidden representation is updated through attention. During the intermediate training stage, a stochastic attention mask blocks direct text/condition-target attention to encourage use of this intermediate pathway.
}
  \label{framework}
\end{fffigure}

\subsection{Preliminaries}
\label{sec:preliminaries}
Modern video generation models typically combine a 3D VAE with a diffusion transformer to model visual content and temporal dynamics in a compressed latent space. We instantiate FrameForge with the first-stage video diffusion model of HunyuanVideo1.5~\cite{hunyuanvideo2025}. Its VAE encoder $\mathcal{E}(\cdot)$ compresses the spatial and temporal dimensions by factors of $16$ and $4$, respectively. Given a video
$\mathbf{x}\in\mathbb{R}^{(4f+1)\times16h\times16w\times c}$,
the encoded latent is
\begin{equation}
    \mathbf{z}_0=\mathcal{E}(\mathbf{x})
    \in\mathbb{R}^{(f+1)\times h\times w\times32}.
\end{equation}

The MM-DiT jointly processes textual and visual tokens, with 3D rotary positional embeddings (3D-RoPE) encoding the temporal and spatial coordinates of visual tokens. The backbone is trained with flow matching~\cite{lipman2022flow}, where the clean latent $\mathbf{z}_0$ is interpolated with Gaussian noise to obtain $\mathbf{z}_t$, and the corresponding velocity $\mathbf{v}$ is predicted by minimizing
\begin{equation}
\mathcal{L}_{\mathrm{diffusion}}
=
\left\|
\mathbf{v}
-
\mathbf{v}_{\theta}(\mathbf{z}_t,t,\mathbf{c})
\right\|_2^2,
\label{eq:flow_matching_loss}
\end{equation}
where $\mathbf{c}$ denotes the conditioning inputs.

\subsection{Framework Overview}
\label{sec:overview}

As shown in \cref{framework}, FrameForge formulates conditional image generation as target-frame prediction guided by textual and visual conditions. The target image is encoded and perturbed, while the visual conditions are encoded into $\mathbf{z}^{\mathrm{cond}}$ using the frozen VAE encoder. In addition, CoTR introduces a noise-initialized reasoning latent $\mathbf{z}^{\mathrm{reason}}$. Together with the text representations $\mathbf{z}^{\mathrm{text}}$, these inputs are jointly processed by the backbone, while MoTE determines their temporal positional configurations. The overall conditioning is expressed as
\begin{equation}
    \mathbf{c}^{\mathrm{ours}}
    =
    \left[
    \mathbf{z}^{\mathrm{text}},
    \mathbf{z}^{\mathrm{cond}},
    \mathbf{z}^{\mathrm{reason}}
    \right].
\end{equation}

\subsection{Mixture of Temporal Experts (MoTE)}
\label{sec:mote}
The 3D-RoPE in the video backbone encodes the temporal coordinates of visual tokens and thereby determines the positional prior for cross-frame interaction. Under conditional frame prediction, using a fixed temporal gap assigns the same positional relation to all condition-target pairs, despite their variations across tasks, samples, and denoising steps. To provide a more flexible temporal prior, we introduce Mixture of Temporal Experts (MoTE), which constructs multiple temporal RoPE experts corresponding to different virtual intervals and uses a lightweight router to select one expert according to the text, condition, and noisy target representations.

\textbf{Temporal RoPE experts.}
We construct \(E\) temporal positional experts with candidate latent spans \(\mathcal{S}=\{s_1,s_2,\ldots,s_E\}\), where \(s_e\) denotes the number of positions along the virtual latent temporal axis for expert \(e\). Assigning the condition and target to the first and last positions yields a virtual temporal interval of \(\Delta_e=s_e-1\). For a causal 3D VAE with temporal compression factor \(r_t\), this latent span corresponds to an equivalent video-frame span of \(F_e=r_t(s_e-1)+1\), including both endpoints. In our main implementation, \(r_t=4\), and the default video-frame spans \(\{17,69,121\}\) correspond to latent spans \(\{5,18,31\}\).

For expert $e$, we first construct the rotary embeddings over a virtual latent grid of size $(s_e,h,w)$:
\begin{equation}
    \left(
    \mathbf{R}^{\cos}_e,
    \mathbf{R}^{\sin}_e
    \right)
    =
    \operatorname{RoPE}
    \left(
    s_e,h,w
    \right).
\end{equation}
For condition-target prediction, we retain the embeddings at temporal coordinates $0$ and $s_e-1$. With CoTR, we additionally retain the embeddings at an intermediate temporal position for the reasoning latent.

\textbf{Sample-wise Top-1 routing.}
For each sample, the router takes the concatenated text, condition, and noisy target representations as input, and produces token-level routing logits:
\begin{equation}
    \mathbf{A}
    =
    G\left(
    \left[
    \mathbf{z}^{\mathrm{text}},
    \mathbf{z}^{\mathrm{cond}},
    \mathbf{z}^{\mathrm{target}}_t
    \right]
    \right)
    \in
    \mathbb{R}^{L\times E},
\end{equation}
where $L$ denotes the routing sequence length and $E$ the number of temporal experts. We then perform masked average pooling over valid tokens to obtain a sample-level logit vector $\mathbf{a}\in\mathbb{R}^{E}$, excluding padded text tokens.

The routing probabilities and Top-1 expert assignment are given by
\begin{equation}
\mathbf{p}
=
\operatorname{Softmax}
\left(
\frac{\mathbf{a}}{\tau_{\mathrm{router}}}
\right),
\qquad
e^{*}
=
\mathop{\arg\max}_{e\in\{1,\ldots,E\}} p_e,
\end{equation}
where $\tau_{\mathrm{router}}$ controls the sharpness of the routing distribution. Moreover, to enable end-to-end optimization with discrete routing, we adopt a straight-through estimator:
\begin{equation}
    \mathbf{g}
    =
    \operatorname{sg}
    \left(
    \mathbf{g}^{\mathrm{hard}}-\mathbf{p}
    \right)
    +
    \mathbf{p},
\end{equation}
where $\mathbf{g}^{\mathrm{hard}}$ is the one-hot gate corresponding to the selected expert $e^{*}$, and $\operatorname{sg}(\cdot)$ denotes the stop-gradient operation. Therefore, the forward pass uses a single selected expert, while gradients are propagated through the soft routing probabilities. At inference time, we directly set
$\mathbf{g}=\mathbf{g}^{\mathrm{hard}}$.

The routed rotary embeddings are then computed as
\begin{equation}
    \mathbf{R}^{\cos}
    =
    \sum_{e=1}^{E}
    g_e\mathbf{R}^{\cos}_e,
    \qquad
    \mathbf{R}^{\sin}
    =
    \sum_{e=1}^{E}
    g_e\mathbf{R}^{\sin}_e,
\end{equation}
and applied to the visual queries and keys in self-attention.

\textbf{Load-balancing regularization.}
To prevent routing collapse and encourage balanced utilization of the temporal experts, we introduce a batch-level load-balancing loss. Let
$\bar{g}_e$
denote the mean gate value of expert $e$ over a batch of size $B$. The balancing loss is defined as
\begin{equation}
    \mathcal{L}_{\mathrm{balance}}
    =
    -\sum_{e=1}^{E}
    \log\left(\bar{g}_e+\epsilon\right),
\end{equation}
where $\epsilon$ is a small constant for numerical stability. 

The overall training objective is
\begin{equation}
    \label{eq:framework_loss}
    \mathcal{L}_{\mathrm{final}}
    =
    \mathcal{L}_{\mathrm{diffusion}}
    +
    \lambda_{\mathrm{bal}}
    \mathcal{L}_{\mathrm{balance}},
\end{equation}
where $\lambda_{\mathrm{bal}}$ controls the strength of regularization.

\subsection{Chain of Temporal Reasoning (CoTR)}
\label{sec:cotr}

\textbf{Intermediate latent representation.}
CoTR introduces an intermediate latent state between the condition and target to model the implicit condition-to-target transformation. Specifically, we insert a single reasoning latent $\mathbf{z}^{\mathrm{reason}}$, sampled from a standard Gaussian distribution with the same spatial and channel dimensions as the visual latents. Rather than being explicitly supervised or decoded, this latent is progressively contextualized through the MM-DiT blocks by interacting with the text, condition, and target representations. It therefore serves as a latent pathway for aggregating conditional information before target prediction. The model is trained solely with the target-frame objective in \cref{eq:framework_loss}, without requiring intermediate supervision.

\textbf{Reasoning-bottleneck attention mask.}
Simply inserting the reasoning latent does not ensure that the model makes effective use of it, since target tokens can still directly attend to the text and condition tokens. To discourage this shortcut, we apply a stochastic reasoning-bottleneck attention mask with probability $p_{\mathrm{mask}}$ during the intermediate training stage.

Let $\mathcal{T}$, $\mathcal{C}$, $\mathcal{R}$, and $\mathcal{X}$ denote the text, condition, reasoning, and target token sets, respectively. The mask $\mathbf{M}^{\mathrm{cotr}}$ is added to the self-attention logits:
\begin{equation}
\mathbf{M}^{\mathrm{cotr}}_{ij}
=
\begin{cases}
-\infty, & i\in\mathcal{T}\cup\mathcal{C},\; j\in\mathcal{X}, \\
-\infty, & i\in\mathcal{X},\; j\in\mathcal{T}\cup\mathcal{C}, \\
0, & \mathrm{otherwise},
\end{cases}
\label{eq:cotr_mask}
\end{equation}
where $i$ and $j$ index query and key tokens, respectively.

The mask removes direct attention between the target and the text/condition tokens while preserving the reasoning latent as a communication bridge between them. Consequently, target tokens receive conditional information through the evolving reasoning representation rather than through a direct attention path. For samples without masking, the original full-attention pattern is retained. This masking only modifies attention connectivity and does not affect the temporal positional conditioning introduced by MoTE.

\subsection{Training Strategy}
\label{sec:lora}

We freeze the pretrained video-backbone weights and the VAE encoder, and insert a single LoRA adapter~\cite{hu2021lora} into the attention and feed-forward projections of the MM-DiT.

\textbf{Three-stage optimization.}
In the first stage, we optimize only the MoTE router with the condition-target formulation, keeping the LoRA parameters fixed and omitting the reasoning latent. This stage learns temporal routing before introducing the additional intermediate pathway. In the second stage, we add the reasoning latent, jointly optimize the router and LoRA, and apply the stochastic mask in \cref{eq:cotr_mask}. In the final stage, we retain both modules and continue joint optimization with full attention. The final stage matches the attention configuration used at inference, where the router selects a temporal expert at each denoising step. Furthermore, the shared adapter is used for every task.
\section{Experiments}
\subsection{Implementation Details}
\textbf{Tasks.} We evaluate FrameForge on three representative conditional image generation tasks: subject-driven generation, canny-to-image generation, and instruction-guided image editing. These tasks respectively assess subject consistency, structural fidelity, and semantic controllability, covering diverse condition-to-target relations within a unified framework.

\textbf{Training data.}
For subject-driven generation, we filter Subject200K~\cite{tan2024ominicontrol} and UNO-1M~\cite{Wu_2025_ICCV} to obtain high-quality reference-target subject pairs with variations in pose, viewpoint, and scene. For canny-to-image generation, we use a subset of 512-2M~\cite{tan2024ominicontrol}, where the extracted edge map serves as the condition and the corresponding image as the target. For image editing, we sample instruction-based editing pairs from the HQ-Edit and UltraEdit subsets of GPT-Image-Edit-1.5M~\cite{wang2025gptimageedit15mmillionscalegptgeneratedimage}. We use 170K samples per task, resulting in 510K training samples in total. All images are resized to $512\times512$.

\textbf{Benchmarks and metrics.}
For subject-driven generation, we evaluate on DreamBench~\cite{ruiz2023dreambooth} using one reference image for each of 30 subjects and 25 prompts per subject. We adopt VIEScore~\cite{ku2023viescore} with GPT-4o and report semantic consistency ($SC$), perceptual quality ($PQ$), and overall score ($O$).

For canny-to-image generation, we use the 5,000-image COCO2017 validation set~\cite{lin2014microsoft}, resized to $512\times512$, with text descriptions generated by GPT-4o. We report FID~\cite{heusel2017gans}, SSIM~\cite{wang2004image}, and CLIP-T~\cite{radford2021learning} for distributional quality, structural fidelity, and text-image alignment, respectively.

For image editing, we evaluate on the English subset of GEdit-Bench~\cite{liu2025step1x-edit}, containing 606 samples across 11 editing categories. Following its official protocol, we report semantic consistency ($G$-$SC$), perceptual quality ($G$-$PQ$), and overall score ($G$-$O$) using GPT-4.1. The $G$- prefix distinguishes GEdit-Bench metrics from the VIEScore metrics used for subject-driven generation.

\textbf{Training and Inference.} Our model, based on HunyuanVideo1.5-T2V-720P and fine-tuned using LoRA with a rank and alpha of 32, has 182M trainable parameters. The model is trained for 12,000 iterations on two NVIDIA A800 GPUs, with a global batch size of 60 and a learning rate of $3\times10^{-4}$, requiring approximately 6 days. For the MoTE router, the softmax temperature is fixed to \(\tau_{\mathrm{router}}=1.0\), and the load-balancing loss is weighted by \(\lambda_{\mathrm{bal}}=0.01\). During the intermediate training stage, we apply the reasoning-bottleneck attention mask with probability $p_{\mathrm{mask}}=0.5$; the final stage and inference use full attention. During inference, we apply classifier-free guidance to the text condition, with the guidance scale $\omega$ set to 6 for subject-driven generation and editing, and to 2 for canny-to-image generation. We use 50 denoising steps for all tasks.
  
\FloatBarrier
\par\ifdim\dimexpr\pagegoal-\pagetotal\relax<8\baselineskip
  \newpage
\fi
\subsection{Main Results}
\subsubsection{Subject-driven image generation}

\paragraph*{Quantitative results}
As shown in \cref{IP-table}, our method achieves the best overall score among all compared methods, reaching an $O$ score of 0.739. It improves over ChronoEdit by 1.37\% and outperforms OminiControl, ACE++, and EasyControl. Its subject consistency of 0.739 ranks second to ChronoEdit. Although EasyControl and IP-Adapter achieve higher perceptual quality scores, their SC scores are substantially lower. Our method thus provides a better balance between subject preservation and image realism.

\begin{fftable}[htbp]
  \centering
\caption{Quantitative comparison on DreamBench for subject-driven image generation}
\label{IP-table}

  { 
      \begin{fftabular}{l l c c c}
        \toprule
\rowcolor{FXPale}
        \textbf{Method} & \textbf{Base Model} & \textbf{SC}$\uparrow$ & \textbf{PQ}$\uparrow$ & \textbf{O}$\uparrow$ \\
        \midrule
        BLIPDiffusion~\cite{li2023blip} & SD 1.5 & 0.574 & 0.696 & 0.613  \\
        SSR-Encoder~\cite{zhang2024ssr} & SD 1.5 & 0.524 & 0.774 & 0.608 \\
        IP-Adapter~\cite{ye2023ip} & SDXL & 0.516 & \underline{0.824} & 0.629 \\
        ACE++~\cite{mao2025ace++} & FLUX.1-dev  & 0.660 & 0.779 & 0.688 \\
        EasyControl~\cite{zhang2025easycontrol} & FLUX.1-dev  & 0.596 & \textbf{0.839} & 0.679 \\
        OmniGen~\cite{xiao2025omnigen} & Native & 0.611 & 0.762 & 0.652 \\
        OminiControl~\cite{tan2024ominicontrol} & FLUX.1-dev & 0.680 & 0.816 & 0.719 \\
        EchoGen~\cite{dong2026echogen} & Infinity-2B & 0.729 & 0.744 & 0.719 \\
        RealGeneral~\cite{lin2025realgeneral} & CogVideoX-1.5 & 0.633 & 0.767 & 0.669 \\
        DRA-Ctrl~\cite{cao2026dimension} & HunyuanVideo-I2V & 0.609 & 0.800 & 0.656 \\
        ChronoEdit~\cite{wu2025chronoedit} & Wan2.1-I2V-14B & \textbf{0.757} & 0.764 & \underline{0.729} \\
        \midrule
        \ffresultrow
Ours & HunyuanVideo1.5 & \underline{0.739} & 0.783 & \textbf{0.739} \\
        \bottomrule
      \end{fftabular}
    }

\end{fftable}

\begin{fffigure}[htbp]
  \centering
  \includegraphics[width=\linewidth,pagebox=cropbox]{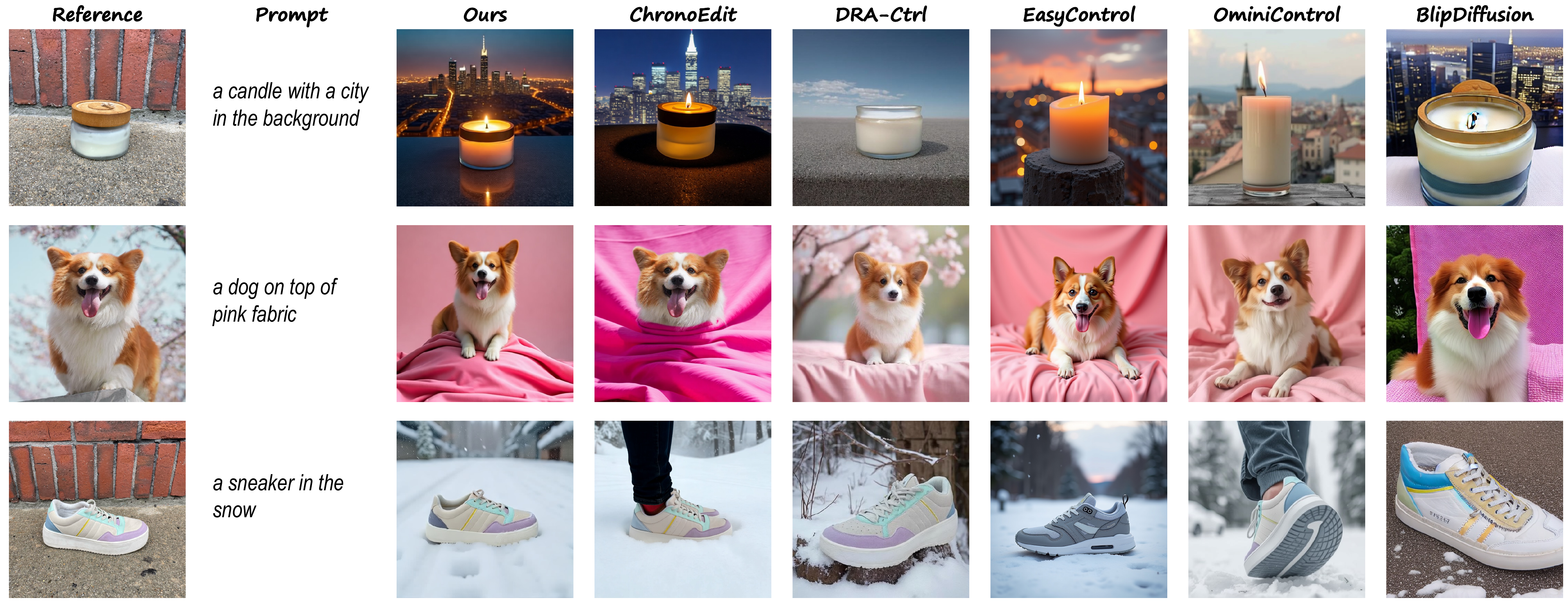}
  \caption{Qualitative comparison on DreamBench for subject-driven image generation. Given a reference subject and a text prompt, our method better preserves instance-level visual attributes while adapting the subject to the prompt-specified scene.}
  \label{ip-case}
\end{fffigure}

\paragraph*{Qualitative results}
\cref{ip-case} illustrates the balance between subject preservation and text alignment. In the dog example, ChronoEdit places the subject partially within the pink fabric, making the requested ``on top of'' relation unclear. EasyControl changes the shoe design in the sneaker example. Our results preserve distinctive attributes while following the requested scenes and spatial relations, including placing the dog on top of the fabric.

\FloatBarrier
\subsubsection{Canny-to-image generation}

\begin{wrapfigure}{r}{.48\textwidth}
  \captionsetup{width=.94\linewidth,position=bottom}
  \centering
  \includegraphics[width=\linewidth]{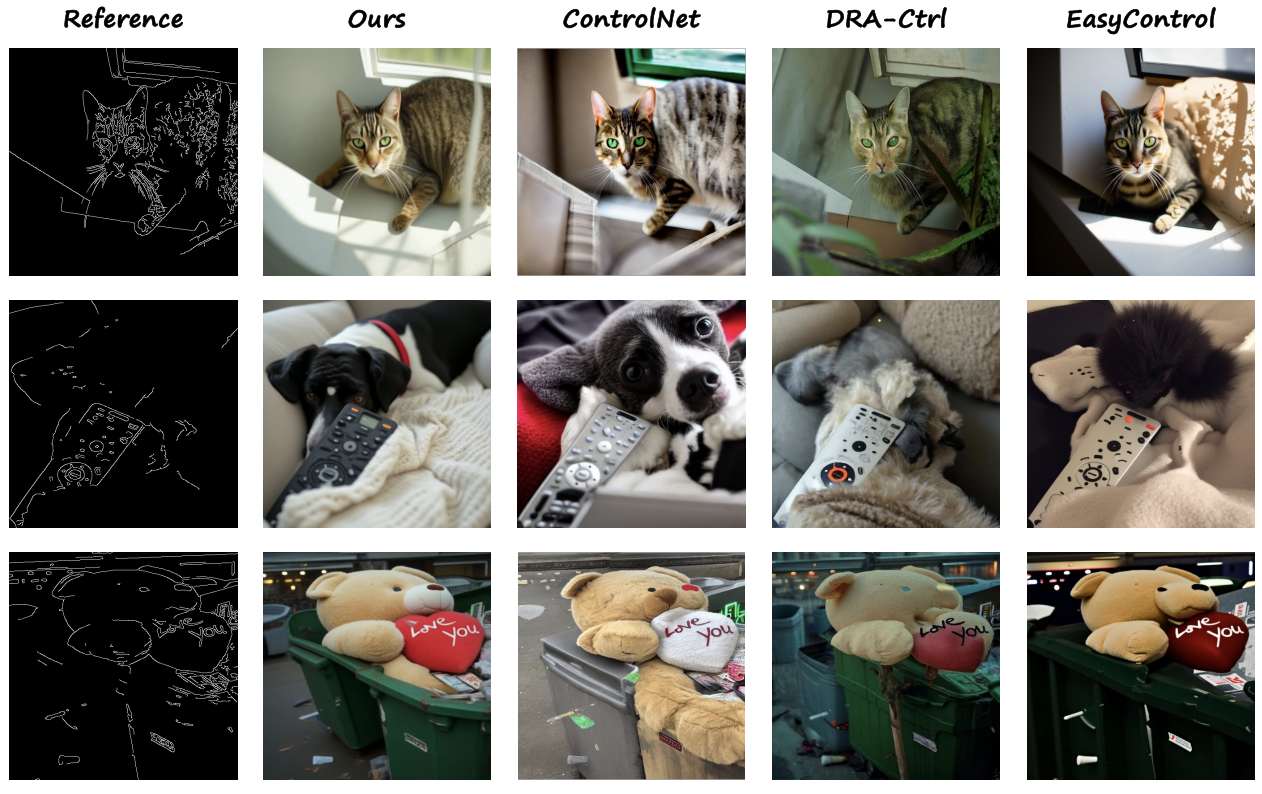}
  \caption{Qualitative comparison on COCO2017 for canny-to-image generation. Our method more faithfully follows the input geometry while producing semantically consistent and visually coherent images.}
  \label{canny-case}
\end{wrapfigure}

\paragraph*{Quantitative results}
As shown in \cref{Canny-table}, FrameForge achieves the highest SSIM of 0.447, corresponding to a relative improvement of 7.2\% over the second-best EasyControl. Its CLIP-T score of 0.323 is comparable to ControlNet and RealGeneral, indicating improved structural fidelity with maintained text-image alignment. Its FID of 23.47 is comparable to those of video-based DRA-Ctrl and RealGeneral.

These metrics capture complementary properties: SSIM measures image-level structural similarity, whereas FID measures distributional differences in Inception features. Backbone-specific appearance statistics may therefore affect FID without necessarily implying inferior perceptual quality. In addition, prior studies also report disagreement with human judgments~\cite{Jayasumana_2024_CVPR,stein2023exposing}. Therefore, we consider FID together with SSIM, CLIP-T, and qualitative results.

\begin{fftable}[htbp]
  \centering
  \caption{Quantitative results on COCO2017 for canny-to-image task}
  \label{Canny-table}
  { 
      \begin{fftabular}{l l c c c}
        \toprule
\rowcolor{FXPale}
        \textbf{Method} & \textbf{Base Model} & \textbf{FID}$\downarrow$ & \textbf{SSIM}$\uparrow$ & \textbf{CLIP-T}$\uparrow$ \\
        \midrule
        ControlNet~\cite{zhang2023adding} & SD1.5 & \textbf{20.42} & 0.299 & 0.323 \\
        OmniGen~\cite{xiao2025omnigen} & Native & \underline{22.51} & 0.352 & \textbf{0.329} \\
        EasyControl~\cite{zhang2025easycontrol} & FLUX.1-dev & 22.82 & \underline{0.417} & 0.319\\
        RealGeneral~\cite{lin2025realgeneral} & CogVideoX-1.5 & 24.63 & 0.414 & \underline{0.324} \\
        DRA-Ctrl~\cite{cao2026dimension} & HunyuanVideo-I2V & 23.27 & 0.408 & 0.314 \\
        \midrule
        \ffresultrow
Ours & HunyuanVideo1.5 & 23.47 & \textbf{0.447} & 0.323 \\
        \bottomrule
      \end{fftabular}
    }
\end{fftable}

\paragraph*{Qualitative results}
In \cref{canny-case}, FrameForge preserves the poses, relative positions, and major contours specified by the edge maps while producing coherent images. Competing methods sometimes alter object geometry or layout. Although ControlNet achieves the lowest FID, its examples show deviations from the input edges, illustrating the complementary roles of distributional and structural metrics.

\FloatBarrier
\subsubsection{Image Editing}
\paragraph*{Quantitative results}
\cref{gedit_results} compares FrameForge with recent instruction-guided image editing methods on GEdit. We report results on both Motion Change and the average over all 11 categories. Existing image-backbone adaptation methods perform poorly on Motion Change, with low $G$-$SC$ and $G$-$O$ scores despite competitive $G$-$PQ$ in some cases. This highlights the importance of the dynamic visual priors learned by video foundation models.

Among open-source methods, FrameForge achieves the best $G$-$O$ on both Motion Change and Overall GEdit, improving over ChronoEdit from 5.76 to 6.84 and from 5.08 to 6.12, respectively. Notably, its Motion Change $G$-$O$ is already comparable to closed-source reference systems, surpassing Gemini 2.5 and Seedream 4.0 while approaching GPT-4o. 

\begin{fftable}[htbp]
\caption{Quantitative comparison on GEdit for instruction-guided image editing}
\label{gedit_results}
\begin{fftabular}{l l c c c c c c}
\toprule
\rowcolor{FXPale}
& & \multicolumn{3}{c}{\textbf{Motion Change}}
  & \multicolumn{3}{c}{\textbf{Overall GEdit}} \\
\rowcolor{FXPale}
\multirow{-2}{*}{\textbf{Method}} & \multirow{-2}{*}{\textbf{Backbone}}
& \textbf{G-SC}$\uparrow$ & \textbf{G-PQ}$\uparrow$ & \textbf{G-O}$\uparrow$
& \textbf{G-SC}$\uparrow$ & \textbf{G-PQ}$\uparrow$ & \textbf{G-O}$\uparrow$ \\
\midrule
\multicolumn{8}{l}{\bfseries Closed-source} \\
\addlinespace[1.5pt]
Gemini 2.5~\cite{gemini25pro2025} & Native & 6.87 & 7.79 & 6.72 & 8.25 & 8.29 & 7.89 \\
GPT-4o~\cite{OpenAI2024} & Native & 7.81 & 8.53 & 7.81 & 8.74 & 7.67 & 8.01 \\
Seedream 4.0~\cite{seedream4web2025} & Native & 5.58 & 8.53 & 5.53 & 8.41 & 8.04 & 7.81 \\
\midrule
\multicolumn{8}{l}{\bfseries Open-source} \\
\addlinespace[1.5pt]
ACE++~\cite{mao2025ace++} & FLUX.1-dev & 1.80 & 6.85 & 2.29 & 2.27 & 6.44 & 2.60 \\
EditAR~\cite{mu2025editar} & LlamaGen & 2.65 & 6.05 & 2.91 & 4.26 & 5.70 & 3.89 \\
ICEdit~\cite{zhang2025icedit} & FLUX.1-dev & 0.93 & \textbf{7.98} & 1.13 & 4.94 & \underline{7.39} & 4.87 \\
OmniGen~\cite{xiao2025omnigen} & Native & 3.35 & 6.68 & 3.12 & \underline{5.88} & 5.87 & 5.01 \\
UniWorld-V1~\cite{lin2025uniworld} & Native & 1.58 & \underline{7.55} & 1.76 & 4.93 & \textbf{7.43} & 4.85 \\
HiDream-I1 (E1)~\cite{cai2025hidream} & Native & 1.58 & 7.23 & 1.66 & 5.66 & 6.06 & 5.01 \\
ChronoEdit~\cite{wu2025chronoedit} & Wan2.1-I2V-14B & \underline{5.90} & 7.33 & \underline{5.76} & 5.35 & 6.88 & \underline{5.08} \\
\midrule
\ffresultrow
\textbf{Ours} & HunyuanVideo1.5 & \textbf{6.90} & 7.18 & \textbf{6.84} & \textbf{6.56} & 6.52 & \textbf{6.12} \\
\bottomrule
\end{fftabular}
\end{fftable}

\paragraph*{Qualitative results}
\cref{edit-case} presents qualitative comparisons across three representative editing types. FrameForge accurately follows the instructions while preserving the source identity and scene layout. Its advantage is particularly evident for motion and spatial edits, such as changing the cat into a sleeping pose and placing the balloon strip below the airplane, where competing methods often leave the requested change incomplete, mislocalize the added content, or introduce content drift.

\begin{fffigure}[htbp]
  \centering
  \includegraphics[width=\linewidth,pagebox=cropbox]{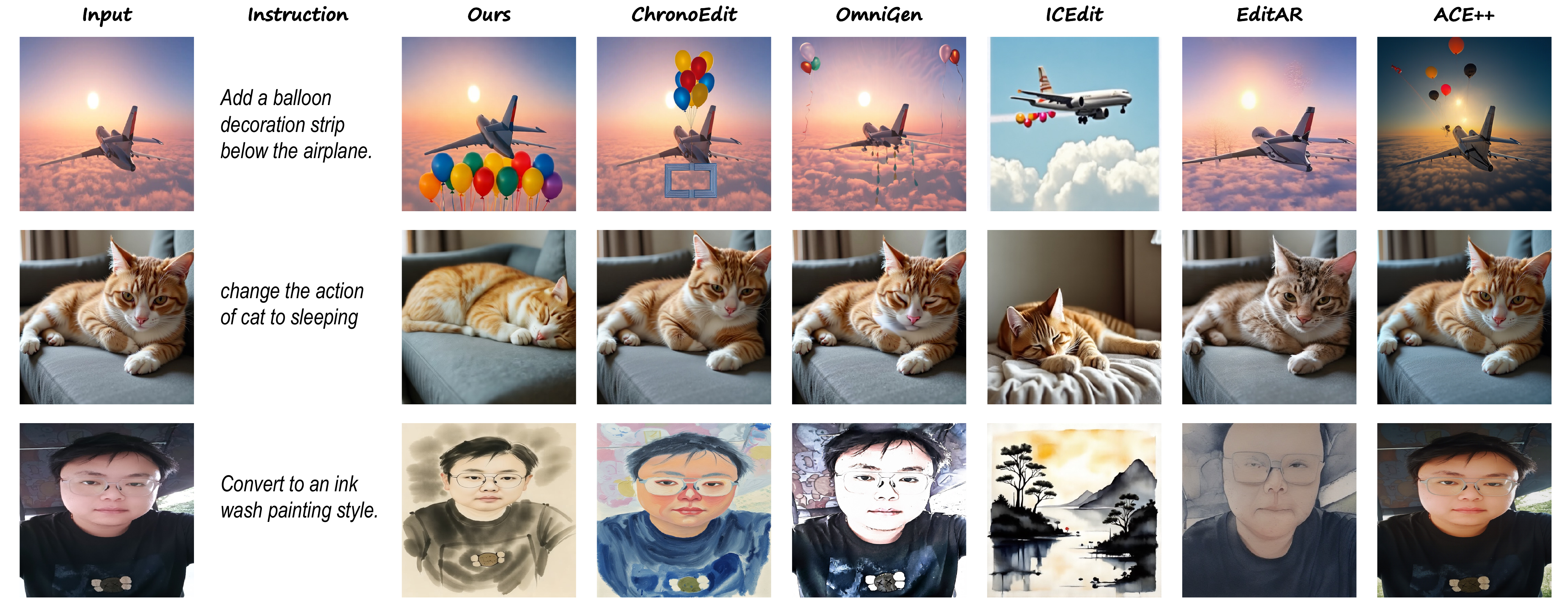}
  \caption{Qualitative comparison of image editing methods. The examples cover object addition, motion change, and style transfer. FrameForge more accurately follows the instructions while preserving the content unrelated to the requested edit.}
  \label{edit-case}
\end{fffigure}

\FloatBarrier
\subsubsection{MoTE Routing Analysis}
\cref{router_vis} reports expert selection ratios over 50 denoising steps, averaged over 500 cases. The MoTE router exhibits distinct task-dependent and timestep-dependent preferences.

\begin{fffigure}[htbp]
  \centering
  \includegraphics[width=\linewidth]{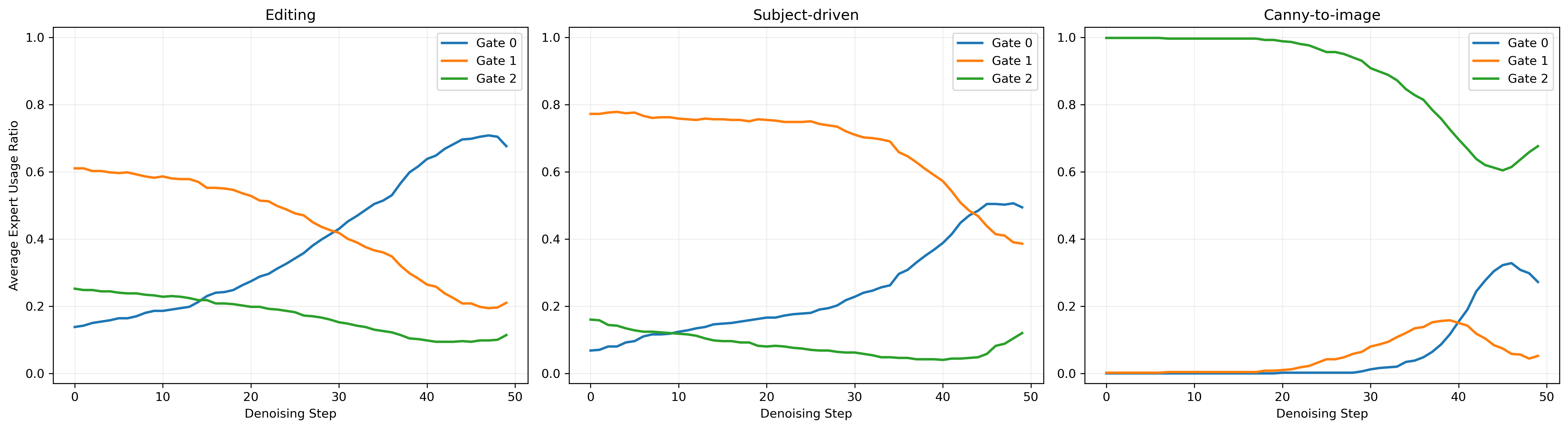}
  \caption{Visualization of MoTE routing behavior for editing, subject-driven generation, and canny-to-image. Gate 0, Gate 1, and Gate 2 denote the short-, medium-, and long-interval temporal RoPE experts, respectively. Each curve reports the average expert usage ratio over 500 cases at each denoising step.}
  \label{router_vis}
\end{fffigure}

Canny-to-image generation, which has the largest condition-target distance, predominantly selects the long-interval expert, particularly during early denoising. Subject-driven generation and editing initially favor the medium-interval expert while target latents remain dominated by noise, with a higher selection ratio for subject-driven generation. As target content emerges, both shift toward the short-interval expert. Editing makes this transition earlier and more strongly, consistent with greater source-content preservation. These within-task shifts suggest that expert selection responds to the evolving target representation, rather than assigning each task a fixed temporal interval. These results complement the dataset-level analysis in \cref{fig:motivation}, showing that MoTE dynamically adapts the virtual temporal relation to both the task and denoising stage.

\subsection{Ablations}
Unless otherwise specified, ablations report overall scores for subject-driven generation and editing (higher is better), and FID for canny-to-image generation (lower is better).

\paragraph*{Main component ablation} Table~\ref{tab:main_ablation} compares task-specific and shared LoRA adaptation. Joint training improves subject-driven generation but degrades editing, suggesting greater task interference for editing. Adding MoTE or CoTR individually mainly improves editing, raising its score from 4.94 to 5.77 or 5.56, respectively, with small changes on the other tasks. When combined, the full model achieves the best overall performance, with the highest scores on subject-driven generation and editing and the lowest FID on canny-to-image generation. This demonstrates that MoTE and CoTR are complementary in adapting the video diffusion backbone to heterogeneous generation tasks.

Notably, neither module alone recovers the task-specific editing score of 5.99, whereas their combination reaches 6.12 and also improves the other two tasks. This is consistent with temporal prior selection and intermediate transformation modeling addressing complementary requirements: adapting only one aspect does not yield the strongest joint performance in this comparison.
 
\begin{fftable}[htbp]
\centering
\caption{Main component ablation}
\label{tab:main_ablation}
\begin{fftabular}{l l c c c c c}
\toprule
\rowcolor{FXPale}
\textbf{Variant} 
& \textbf{LoRA} 
& \textbf{MoTE} 
& \textbf{CoTR} 
& \textbf{Subject $\uparrow$} 
& \textbf{Canny $\downarrow$} 
& \textbf{Editing $\uparrow$} \\
\midrule
Task-specific baseline 
& Task-specific & \xmark & \xmark 
& 0.676 & 25.22 & 5.99 \\
Unified baseline 
& Shared & \xmark & \xmark 
& 0.722 & 25.78 & 4.94 \\
+ MoTE 
& Shared & \cmark & \xmark 
& 0.719 & 25.62 & 5.77 \\
+ CoTR
& Shared & \xmark & \cmark 
& 0.726 & 25.94 & 5.56 \\
\ffresultrow
Full Model 
& Shared & \cmark & \cmark 
& \textbf{0.739} & \textbf{23.47} & \textbf{6.12} \\
\bottomrule
\end{fftabular}
\fftablenote{Here, we compare task-specific LoRA with unified shared LoRA and evaluate the contributions of MoTE and CoTR.}
\end{fftable}

\paragraph*{Number of experts}
As shown in \cref{tab:num_experts}, increasing the expert count from one to three improves canny-to-image generation and editing while maintaining the subject-driven score. Five or seven experts slightly improve subject-driven generation but degrade the other tasks. A larger pool may distribute supervision too sparsely across experts under joint training. Therefore, we use three temporal RoPE experts for the best overall trade-off.

\begin{fftable}[htbp]
\centering
\caption{Ablation on the number of temporal RoPE experts in MoTE}
\label{tab:num_experts}
\begin{fftabular}{c c c c}
\toprule
\rowcolor{FXPale}
\textbf{\# Experts} 
& \textbf{Subject $\uparrow$} 
& \textbf{Canny $\downarrow$} 
& \textbf{Editing $\uparrow$} \\
\midrule
1 & 0.739 & 25.90 & 5.90 \\
\ffresultrow
3 & 0.739 & \textbf{23.47} & \textbf{6.12} \\
5 & \textbf{0.743} & 24.24 & 5.96 \\
7 & \textbf{0.743} & 24.55 & 5.67 \\
\bottomrule
\end{fftabular}
\fftablenote{CoTR is enabled and other settings are fixed. One expert uses virtual video-frame indices 1 and 121. The CoTR-only variant in Table~\ref{tab:main_ablation} uses indices 1 and 17.}
\end{fftable}

\paragraph*{Virtual temporal intervals}
\cref{tab:expert_distance} compares candidate intervals through their equivalent video-frame spans ($F_e$). Short-range experts achieve the highest subject-driven score, while medium-range experts achieve the best canny FID but degrade the other tasks. The wide-range configuration achieves the best editing performance, while remaining comparable to the best results on subject-driven and canny-to-image generation. Its subject-driven score is only 0.002 below the best, supporting the use of $\{17,69,121\}$ to cover diverse condition-target relations.

\begin{fftable}[htbp]
\centering
\caption{Ablation on virtual temporal interval configurations}
\label{tab:expert_distance}
{
\begin{fftabular}{l c c c c}
\toprule
\rowcolor{FXPale}
\textbf{Setting} & \textbf{Video Spans} 
& 
\textbf{Subject $\uparrow$} 
& \textbf{Canny $\downarrow$} 
& \textbf{Editing $\uparrow$} \\
\midrule
Short & $\{17, 25, 33\}$ & \textbf{0.741} & 24.84 & 5.71 \\
Medium & $\{13, 37, 61\}$ & 0.716 & \textbf{23.19} & 5.42 \\
\ffresultrow
Long & $\{17, 69, 121\}$ & 0.739 & 23.47 & \textbf{6.12} \\
\bottomrule
\end{fftabular}
}
\fftablenote{We fix the number of experts and vary their equivalent virtual video spans.}
\end{fftable}

\paragraph*{Number of reasoning latents}
\cref{tab:num_reasoning_latents} shows that introducing one reasoning latent improves all three tasks compared with using none. Two or three latents reduce performance, especially on subject-driven generation and editing. Without direct intermediate supervision, additional latents may enlarge the under-constrained transformation space and disperse the target training signal. The results indicate that one reasoning latent already provides sufficient transformation capacity while retaining an effective information bottleneck. We therefore use a single reasoning latent by default.

\begin{fftable}[htbp]
\centering
\caption{Ablation on the number of reasoning latents in CoTR}
\label{tab:num_reasoning_latents}
{
\begin{fftabular}{c c c c}
\toprule
\rowcolor{FXPale}
\textbf{\# Reasoning Latents} 
& \textbf{Subject $\uparrow$} 
& \textbf{Canny $\downarrow$} 
& \textbf{Editing $\uparrow$} \\
\midrule
0 & 0.719 & 25.62 & 5.77 \\
\ffresultrow
1 & \textbf{0.739} & \textbf{23.47} & \textbf{6.12} \\
2 & 0.709 & 25.82 & 5.62 \\
3 & 0.713 & 24.55 & 5.09 \\
\bottomrule
\end{fftabular}
}
\fftablenote{CoTR uses one reasoning latent by default.}
\end{fftable}

\subsection{Generalization Across Video Backbones}
\label{sec:backbone_generalization}
We further evaluate FrameForge on HunyuanVideo1.0 against a baseline trained with a shared LoRA across the three tasks. We report overall scores for subject-driven generation and editing, and FID for canny-to-image generation.

As shown in \cref{tab:backbone_generalization}, FrameForge consistently outperforms the unified baseline across all three tasks. It raises the subject-driven score from 0.699 to 0.736 and the editing score from 4.38 to 5.00, while reducing canny FID from 24.52 to 23.07. These improvements support the applicability of the proposed adaptation strategy beyond the main backbone.

\begin{fftable}[htbp]
\centering
\caption{Evaluation on HunyuanVideo1.0}
\label{tab:backbone_generalization}
\begin{fftabular}{l c c c}
\toprule
\rowcolor{FXPale}
\textbf{Method} & \textbf{Subject $\uparrow$} & \textbf{Canny $\downarrow$}
       & \textbf{Editing $\uparrow$} \\
\midrule
Unified baseline & 0.699 &  24.52     & 4.38 \\
\ffresultrow
FrameForge       & \textbf{0.736} & \textbf{23.07}
                 & \textbf{5.00} \\
\bottomrule
\end{fftabular}
\end{fftable}

\section{Conclusion}
In this work, we presented FrameForge, a unified adaptation framework that leverages the temporal priors of pretrained video foundation models for heterogeneous conditional image generation. Rather than assigning a fixed temporal relation to all condition-target pairs, we introduced an input-adaptive virtual temporal relation perspective. Based on this formulation, MoTE selects an appropriate temporal positional prior from multiple interval experts according to the textual and visual context. CoTR introduces an unsupervised reasoning latent and a reasoning-bottleneck attention mask to model the unobserved condition-to-target transformation. The proposed framework enables joint adaptation across subject-driven generation, canny-to-image generation, and image editing while keeping the video foundation model frozen. Experiments on DreamBench, COCO2017, and GEdit demonstrate strong performance across all three tasks, and the ablation studies verify the effectiveness of both adaptive temporal relation modeling and implicit temporal reasoning.

\begingroup\small
\bibliographystyle{plainnat}
\bibliography{egbib}

\begin{thebibliography}{67}
\providecommand{\natexlab}[1]{#1}
\providecommand{\url}[1]{\texttt{#1}}
\expandafter\ifx\csname urlstyle\endcsname\relax
  \providecommand{\doi}[1]{doi: #1}\else
  \providecommand{\doi}{doi: \begingroup \urlstyle{rm}\Url}\fi

\bibitem[Acuaviva et~al.(2025)Acuaviva, Davtyan, Hassan, Stapf, Rahimi, Alahi,
  and Favaro]{acuaviva2025rethinking}
Pablo Acuaviva, Aram Davtyan, Mariam Hassan, Sebastian Stapf, Ahmad Rahimi,
  Alexandre Alahi, and Paolo Favaro.
\newblock Rethinking visual intelligence: Insights from video pretraining.
\newblock In \emph{Forty-third International Conference on Machine Learning},
  2025.

\bibitem[Bai et~al.(2024)Bai, Geng, Mangalam, Bar, Yuille, Darrell, Malik, and
  Efros]{bai2024sequential}
Yutong Bai, Xinyang Geng, Karttikeya Mangalam, Amir Bar, Alan~L Yuille, Trevor
  Darrell, Jitendra Malik, and Alexei~A Efros.
\newblock Sequential modeling enables scalable learning for large vision
  models.
\newblock In \emph{Proceedings of the IEEE/CVF Conference on Computer Vision
  and Pattern Recognition}, pages 22861--22872, 2024.

\bibitem[Brooks et~al.(2023)Brooks, Holynski, and
  Efros]{brooks2023instructpix2pix}
Tim Brooks, Aleksander Holynski, and Alexei~A Efros.
\newblock Instructpix2pix: Learning to follow image editing instructions.
\newblock In \emph{Proceedings of the IEEE/CVF conference on computer vision
  and pattern recognition}, pages 18392--18402, 2023.

\bibitem[{ByteDance Seed}(2025)]{seedream4web2025}
{ByteDance Seed}.
\newblock Seedream 4.0.
\newblock \url{https://seed.bytedance.com/en/seedream4_0}, 2025.

\bibitem[Cai et~al.(2025)Cai, Chen, Chen, Li, Long, Pan, Qiu, Zhang, Gao, Xu,
  et~al.]{cai2025hidream}
Qi~Cai, Jingwen Chen, Yang Chen, Yehao Li, Fuchen Long, Yingwei Pan, Zhaofan
  Qiu, Yiheng Zhang, Fengbin Gao, Peihan Xu, et~al.
\newblock Hidream-i1: A high-efficient image generative foundation model with
  sparse diffusion transformer.
\newblock \emph{arXiv preprint arXiv:2505.22705}, 2025.

\bibitem[Cao et~al.(2026)Cao, Feng, Gong, Tian, Lu, Liu, and
  Wang]{cao2026dimension}
Hengyuan Cao, Yutong Feng, Biao Gong, Yijing Tian, Yunhong Lu, Chuang Liu, and
  Bin Wang.
\newblock Dimension-reduction attack! video generative models are experts on
  controllable image synthesis.
\newblock \emph{Advances in Neural Information Processing Systems},
  38:\penalty0 81329--81358, 2026.

\bibitem[Chen et~al.(2025)Chen, Zhao, Yu, Chu, Chen, Yang, Wang, Pan, Zhou,
  Ling, et~al.]{chen2025sana}
Junsong Chen, Yuyang Zhao, Jincheng Yu, Ruihang Chu, Junyu Chen, Shuai Yang,
  Xianbang Wang, Yicheng Pan, Daquan Zhou, Huan Ling, et~al.
\newblock Sana-video: Efficient video generation with block linear diffusion
  transformer.
\newblock \emph{arXiv preprint arXiv:2509.24695}, 2025.

\bibitem[Chen et~al.(2026)Chen, Wang, Pan, Zhu, Yang, and
  Zhang]{chen2026detail++}
Lifeng Chen, Jiner Wang, Zihao Pan, Beier Zhu, Xiaofeng Yang, and Chi Zhang.
\newblock Detail++: Training-free detail enhancer for t2i diffusion models.
\newblock \emph{IEEE Transactions on Image Processing}, 2026.

\bibitem[Chen et~al.(2024)Chen, Zhang, Zhang, Zhou, Kim, Liu, Li, Zhang, Zhao,
  Wang, et~al.]{chen2024unireal}
Xi~Chen, Zhifei Zhang, He~Zhang, Yuqian Zhou, Soo~Ye Kim, Qing Liu, Yijun Li,
  Jianming Zhang, Nanxuan Zhao, Yilin Wang, et~al.
\newblock Unireal: Universal image generation and editing via learning
  real-world dynamics.
\newblock \emph{arXiv preprint arXiv:2412.07774}, 2024.

\bibitem[DeepMind(2025)]{gemini25pro2025}
Google DeepMind.
\newblock Gemini 2.5.
\newblock \url{https://deepmind.google/technologies/gemini/}, 2025.

\bibitem[Deng et~al.(2025)Deng, Zhu, Li, Gou, Li, Wang, Zhong, Yu, Nie, Song,
  Shi, and Fan]{deng2025bagel}
Chaorui Deng, Deyao Zhu, Kunchang Li, Chenhui Gou, Feng Li, Zeyu Wang, Shu
  Zhong, Weihao Yu, Xiaonan Nie, Ziang Song, Guang Shi, and Haoqi Fan.
\newblock Emerging properties in unified multimodal pretraining.
\newblock \emph{arXiv preprint arXiv:2505.14683}, 2025.

\bibitem[Dong et~al.(2026)Dong, Wang, Liu, Li, Chen, Li, Li, and
  Li]{dong2026echogen}
Ruixiao Dong, Zhendong Wang, Keli Liu, Li~Li, Ying Chen, Kai Li, Daowen Li, and
  Houqiang Li.
\newblock Echogen: Generating visual echoes in any scene via feed-forward
  subject-driven auto-regressive model.
\newblock In \emph{The Fourteenth International Conference on Learning
  Representations}, 2026.
\newblock URL \url{https://openreview.net/forum?id=ctmyCjo18u}.

\bibitem[Esser et~al.(2021)Esser, Rombach, and Ommer]{esser2021taming}
Patrick Esser, Robin Rombach, and Bjorn Ommer.
\newblock Taming transformers for high-resolution image synthesis.
\newblock In \emph{Proceedings of the IEEE/CVF conference on computer vision
  and pattern recognition}, pages 12873--12883, 2021.

\bibitem[Esser et~al.(2024)Esser, Kulal, Blattmann, Entezari, Müller, Saini,
  Levi, Lorenz, Sauer, Boesel, Podell, Dockhorn, English, Lacey, Goodwin,
  Marek, and Rombach]{esser2024scalingrectifiedflowtransformers}
Patrick Esser, Sumith Kulal, Andreas Blattmann, Rahim Entezari, Jonas Müller,
  Harry Saini, Yam Levi, Dominik Lorenz, Axel Sauer, Frederic Boesel, Dustin
  Podell, Tim Dockhorn, Zion English, Kyle Lacey, Alex Goodwin, Yannik Marek,
  and Robin Rombach.
\newblock Scaling rectified flow transformers for high-resolution image
  synthesis, 2024.
\newblock URL \url{https://arxiv.org/abs/2403.03206}.

\bibitem[Fu et~al.(2026)Fu, Huang, Wu, Jiang, Huo, Li, Song, Ding, Guo, He, Fu,
  Mao, and Zhang]{fu2026lanceunifiedmultimodalmodeling}
Fengyi Fu, Mengqi Huang, Shaojin Wu, Yunsheng Jiang, Yufei Huo, Hao Li,
  Yinghang Song, Fei Ding, Jianzhu Guo, Qian He, Zheren Fu, Zhendong Mao, and
  Yongdong Zhang.
\newblock Lance: Unified multimodal modeling by multi-task synergy, 2026.
\newblock URL \url{https://arxiv.org/abs/2605.18678}.

\bibitem[Fu et~al.(2025)Fu, Zeng, Lan, Liao, Chen, Chen, Wei, Cheng, Liu, Chen,
  Yu, and Lin]{fu2025iMontage}
Zhoujie Fu, Xianfang Zeng, Jinghong Lan, Xinyao Liao, Cheng Chen, Junyi Chen,
  Jiacheng Wei, Wei Cheng, Shiyu Liu, Yunuo Chen, Gang Yu, and Guosheng Lin.
\newblock imontage: Unified, versatile, highly dynamic many-to-many image
  generation.
\newblock \emph{arXiv preprint arXiv:2511.20635}, 2025.

\bibitem[Guo et~al.(2024)Guo, Yang, Rao, Liang, Wang, Qiao, Agrawala, Lin, and
  Dai]{guo2023animatediff}
Yuwei Guo, Ceyuan Yang, Anyi Rao, Zhengyang Liang, Yaohui Wang, Yu~Qiao,
  Maneesh Agrawala, Dahua Lin, and Bo~Dai.
\newblock Animatediff: Animate your personalized text-to-image diffusion models
  without specific tuning.
\newblock \emph{International Conference on Learning Representations}, 2024.

\bibitem[Heusel et~al.(2017)Heusel, Ramsauer, Unterthiner, Nessler, and
  Hochreiter]{heusel2017gans}
Martin Heusel, Hubert Ramsauer, Thomas Unterthiner, Bernhard Nessler, and Sepp
  Hochreiter.
\newblock Gans trained by a two time-scale update rule converge to a local nash
  equilibrium.
\newblock \emph{Advances in neural information processing systems}, 30, 2017.

\bibitem[Ho et~al.(2020)Ho, Jain, and Abbeel]{ho2020denoising}
Jonathan Ho, Ajay Jain, and Pieter Abbeel.
\newblock Denoising diffusion probabilistic models.
\newblock \emph{Advances in neural information processing systems},
  33:\penalty0 6840--6851, 2020.

\bibitem[Hu et~al.(2021)Hu, Shen, Wallis, Allen-Zhu, Li, Wang, Wang, and
  Chen]{hu2021lora}
Edward~J Hu, Yelong Shen, Phillip Wallis, Zeyuan Allen-Zhu, Yuanzhi Li, Shean
  Wang, Lu~Wang, and Weizhu Chen.
\newblock Lora: Low-rank adaptation of large language models.
\newblock \emph{arXiv preprint arXiv:2106.09685}, 2021.

\bibitem[Huang et~al.(2023)Huang, Mao, Chen, and Zhang]{huang2023towards}
Mengqi Huang, Zhendong Mao, Zhuowei Chen, and Yongdong Zhang.
\newblock Towards accurate image coding: Improved autoregressive image
  generation with dynamic vector quantization.
\newblock In \emph{Proceedings of the IEEE/CVF Conference on Computer Vision
  and Pattern Recognition}, pages 22596--22605, 2023.

\bibitem[Jayasumana et~al.(2024)Jayasumana, Ramalingam, Veit, Glasner,
  Chakrabarti, and Kumar]{Jayasumana_2024_CVPR}
Sadeep Jayasumana, Srikumar Ramalingam, Andreas Veit, Daniel Glasner, Ayan
  Chakrabarti, and Sanjiv Kumar.
\newblock Rethinking fid: Towards a better evaluation metric for image
  generation.
\newblock In \emph{Proceedings of the IEEE/CVF Conference on Computer Vision
  and Pattern Recognition (CVPR)}, pages 9307--9315, June 2024.

\bibitem[Kong et~al.(2024)Kong, Tian, Zhang, Min, Dai, Zhou, Xiong, Li, Wu,
  Zhang, et~al.]{kong2024hunyuanvideo}
Weijie Kong, Qi~Tian, Zijian Zhang, Rox Min, Zuozhuo Dai, Jin Zhou, Jiangfeng
  Xiong, Xin Li, Bo~Wu, Jianwei Zhang, et~al.
\newblock Hunyuanvideo: A systematic framework for large video generative
  models.
\newblock \emph{arXiv preprint arXiv:2412.03603}, 2024.

\bibitem[Ku et~al.(2023)Ku, Jiang, Wei, Yue, and Chen]{ku2023viescore}
Max Ku, Dongfu Jiang, Cong Wei, Xiang Yue, and Wenhu Chen.
\newblock Viescore: Towards explainable metrics for conditional image synthesis
  evaluation, 2023.

\bibitem[Labs(2024)]{flux2024}
Black~Forest Labs.
\newblock Flux: Official inference repository for flux.1 models.
\newblock \url{https://github.com/black-forest-labs/flux}, 2024.
\newblock Accessed: 2024-11-12.

\bibitem[Li et~al.(2023)Li, Li, and Hoi]{li2023blip}
Dongxu Li, Junnan Li, and Steven Hoi.
\newblock Blip-diffusion: Pre-trained subject representation for controllable
  text-to-image generation and editing.
\newblock \emph{Advances in Neural Information Processing Systems},
  36:\penalty0 30146--30166, 2023.

\bibitem[Li et~al.(2025)Li, Sun, Wu, Duan, Wang, Bo, Zhang, and
  Liang]{li2025video4editviewingimageediting}
Xiaofan Li, Yanpeng Sun, Chenming Wu, Fan Duan, YuAn Wang, Weihao Bo, Yumeng
  Zhang, and Dingkang Liang.
\newblock Video4edit: Viewing image editing as a degenerate temporal process,
  2025.
\newblock URL \url{https://arxiv.org/abs/2511.18131}.

\bibitem[Lin et~al.(2025{\natexlab{a}})Lin, Li, Cheng, Niu, Ye, He, Yuan, Yu,
  Wang, Ge, et~al.]{lin2025uniworld}
Bin Lin, Zongjian Li, Xinhua Cheng, Yuwei Niu, Yang Ye, Xianyi He, Shenghai
  Yuan, Wangbo Yu, Shaodong Wang, Yunyang Ge, et~al.
\newblock Uniworld-v1: High-resolution semantic encoders for unified visual
  understanding and generation.
\newblock \emph{arXiv preprint arXiv:2506.03147}, 2025{\natexlab{a}}.

\bibitem[Lin et~al.(2014)Lin, Maire, Belongie, Hays, Perona, Ramanan,
  Doll{\'a}r, and Zitnick]{lin2014microsoft}
Tsung-Yi Lin, Michael Maire, Serge Belongie, James Hays, Pietro Perona, Deva
  Ramanan, Piotr Doll{\'a}r, and C~Lawrence Zitnick.
\newblock Microsoft coco: Common objects in context.
\newblock In \emph{Computer Vision--ECCV 2014: 13th European Conference,
  Zurich, Switzerland, September 6-12, 2014, Proceedings, Part V 13}, pages
  740--755. Springer, 2014.

\bibitem[Lin et~al.(2024)Lin, Wei, Zhang, Zhuo, Zhao, Huang, Xie, Qiao, Gao,
  and Li]{lin2024pixwizard}
Weifeng Lin, Xinyu Wei, Renrui Zhang, Le~Zhuo, Shitian Zhao, Siyuan Huang,
  Junlin Xie, Yu~Qiao, Peng Gao, and Hongsheng Li.
\newblock Pixwizard: Versatile image-to-image visual assistant with
  open-language instructions.
\newblock \emph{arXiv preprint arXiv:2409.15278}, 2024.

\bibitem[Lin et~al.(2025{\natexlab{b}})Lin, Huang, Zhuang, and
  Mao]{lin2025realgeneral}
Yijing Lin, Mengqi Huang, Shuhan Zhuang, and Zhendong Mao.
\newblock Realgeneral: Unifying visual generation via temporal in-context
  learning with video models.
\newblock In \emph{Proceedings of the IEEE/CVF International Conference on
  Computer Vision}, pages 14994--15004, 2025{\natexlab{b}}.

\bibitem[Lipman et~al.(2022)Lipman, Chen, Ben-Hamu, Nickel, and
  Le]{lipman2022flow}
Yaron Lipman, Ricky~TQ Chen, Heli Ben-Hamu, Maximilian Nickel, and Matt Le.
\newblock Flow matching for generative modeling.
\newblock \emph{arXiv preprint arXiv:2210.02747}, 2022.

\bibitem[Liu et~al.(2026)Liu, Li, Zhang, Lan, Luo, and Liu]{liu2026lacon}
Chang Liu, Rui Li, Kaidong Zhang, Yunwei Lan, Xin Luo, and Dong Liu.
\newblock Lacon: Late-constraint controllable visual generation.
\newblock \emph{IEEE Transactions on Image Processing}, 2026.

\bibitem[Liu et~al.(2025{\natexlab{a}})Liu, Han, Xing, Yin, Wang, Cheng, Liao,
  Wang, Fu, Han, Li, Peng, Sun, Wu, Cai, Ge, Ming, Xia, Zeng, Zhu, Jiao, Zhang,
  Yu, and Jiang]{liu2025step1x-edit}
Shiyu Liu, Yucheng Han, Peng Xing, Fukun Yin, Rui Wang, Wei Cheng, Jiaqi Liao,
  Yingming Wang, Honghao Fu, Chunrui Han, Guopeng Li, Yuang Peng, Quan Sun,
  Jingwei Wu, Yan Cai, Zheng Ge, Ranchen Ming, Lei Xia, Xianfang Zeng, Yibo
  Zhu, Binxing Jiao, Xiangyu Zhang, Gang Yu, and Daxin Jiang.
\newblock Step1x-edit: A practical framework for general image editing.
\newblock \emph{arXiv preprint arXiv:2504.17761}, 2025{\natexlab{a}}.

\bibitem[Liu et~al.(2025{\natexlab{b}})Liu, Wang, Li, van~de Weijer, Khan,
  Yang, Wang, Yang, and Cheng]{liu2025one}
Tao Liu, Kai Wang, Senmao Li, Joost van~de Weijer, Fahad~Shahbaz Khan, Shiqi
  Yang, Yaxing Wang, Jian Yang, and Ming-Ming Cheng.
\newblock One-prompt-one-story: Free-lunch consistent text-to-image generation
  using a single prompt.
\newblock \emph{arXiv preprint arXiv:2501.13554}, 2025{\natexlab{b}}.

\bibitem[Ma et~al.(2024)Ma, Wang, Jia, Chen, Liu, Li, Chen, and
  Qiao]{ma2024latte}
Xin Ma, Yaohui Wang, Gengyun Jia, Xinyuan Chen, Ziwei Liu, Yuan-Fang Li,
  Cunjian Chen, and Yu~Qiao.
\newblock Latte: Latent diffusion transformer for video generation.
\newblock \emph{arXiv preprint arXiv:2401.03048}, 2024.

\bibitem[Mao et~al.(2025)Mao, Zhang, Pan, Jiang, Han, Liu, and
  Zhou]{mao2025ace++}
Chaojie Mao, Jingfeng Zhang, Yulin Pan, Zeyinzi Jiang, Zhen Han, Yu~Liu, and
  Jingren Zhou.
\newblock Ace++: Instruction-based image creation and editing via context-aware
  content filling.
\newblock \emph{arXiv preprint arXiv:2501.02487}, 2025.

\bibitem[Mou et~al.(2024)Mou, Wang, Xie, Wu, Zhang, Qi, and Shan]{mou2024t2i}
Chong Mou, Xintao Wang, Liangbin Xie, Yanze Wu, Jian Zhang, Zhongang Qi, and
  Ying Shan.
\newblock T2i-adapter: Learning adapters to dig out more controllable ability
  for text-to-image diffusion models.
\newblock In \emph{Proceedings of the AAAI conference on artificial
  intelligence}, volume~38, pages 4296--4304, 2024.

\bibitem[Mu et~al.(2025)Mu, Vasconcelos, and Wang]{mu2025editar}
Jiteng Mu, Nuno Vasconcelos, and Xiaolong Wang.
\newblock Editar: Unified conditional generation with autoregressive models.
\newblock In \emph{Proceedings of the Computer Vision and Pattern Recognition
  Conference}, pages 7899--7909, 2025.

\bibitem[OpenAI(2024)]{OpenAI2024}
OpenAI.
\newblock Gpt-4o system card, 2024.
\newblock URL \url{https://arxiv.org/html/2410.21276v1}.

\bibitem[Podell et~al.(2023)Podell, English, Lacey, Blattmann, Dockhorn,
  M{\"u}ller, Penna, and Rombach]{podell2023sdxl}
Dustin Podell, Zion English, Kyle Lacey, Andreas Blattmann, Tim Dockhorn, Jonas
  M{\"u}ller, Joe Penna, and Robin Rombach.
\newblock Sdxl: Improving latent diffusion models for high-resolution image
  synthesis.
\newblock \emph{arXiv preprint arXiv:2307.01952}, 2023.

\bibitem[Qin et~al.(2023)Qin, Zhang, Yu, Feng, Yang, Zhou, Wang, Niebles,
  Xiong, Savarese, et~al.]{qin2023unicontrol}
Can Qin, Shu Zhang, Ning Yu, Yihao Feng, Xinyi Yang, Yingbo Zhou, Huan Wang,
  Juan~Carlos Niebles, Caiming Xiong, Silvio Savarese, et~al.
\newblock Unicontrol: A unified diffusion model for controllable visual
  generation in the wild.
\newblock \emph{arXiv preprint arXiv:2305.11147}, 2023.

\bibitem[Radford et~al.(2021)Radford, Kim, Hallacy, Ramesh, Goh, Agarwal,
  Sastry, Askell, Mishkin, Clark, et~al.]{radford2021learning}
Alec Radford, Jong~Wook Kim, Chris Hallacy, Aditya Ramesh, Gabriel Goh,
  Sandhini Agarwal, Girish Sastry, Amanda Askell, Pamela Mishkin, Jack Clark,
  et~al.
\newblock Learning transferable visual models from natural language
  supervision.
\newblock In \emph{International conference on machine learning}, pages
  8748--8763. PMLR, 2021.

\bibitem[Ruiz et~al.(2023)Ruiz, Li, Jampani, Pritch, Rubinstein, and
  Aberman]{ruiz2023dreambooth}
Nataniel Ruiz, Yuanzhen Li, Varun Jampani, Yael Pritch, Michael Rubinstein, and
  Kfir Aberman.
\newblock Dreambooth: Fine tuning text-to-image diffusion models for
  subject-driven generation.
\newblock In \emph{Proceedings of the IEEE/CVF conference on computer vision
  and pattern recognition}, pages 22500--22510, 2023.

\bibitem[Stein et~al.(2023)Stein, Cresswell, Hosseinzadeh, Sui, Ross,
  Villecroze, Liu, Caterini, Taylor, and Loaiza-Ganem]{stein2023exposing}
George Stein, Jesse Cresswell, Rasa Hosseinzadeh, Yi~Sui, Brendan Ross,
  Valentin Villecroze, Zhaoyan Liu, Anthony~L Caterini, Eric Taylor, and
  Gabriel Loaiza-Ganem.
\newblock Exposing flaws of generative model evaluation metrics and their
  unfair treatment of diffusion models.
\newblock \emph{Advances in Neural Information Processing Systems},
  36:\penalty0 3732--3784, 2023.

\bibitem[Sun et~al.(2024)Sun, Chu, Zhang, Wu, Dong, Zang, Xiong, Lin, and
  Wang]{sun2024x}
Zeyi Sun, Ziyang Chu, Pan Zhang, Tong Wu, Xiaoyi Dong, Yuhang Zang, Yuanjun
  Xiong, Dahua Lin, and Jiaqi Wang.
\newblock X-prompt: Towards universal in-context image generation in
  auto-regressive vision language foundation models.
\newblock \emph{arXiv preprint arXiv:2412.01824}, 2024.

\bibitem[Tan et~al.(2024)Tan, Liu, Yang, Xue, and Wang]{tan2024ominicontrol}
Zhenxiong Tan, Songhua Liu, Xingyi Yang, Qiaochu Xue, and Xinchao Wang.
\newblock Ominicontrol: Minimal and universal control for diffusion
  transformer.
\newblock \emph{arXiv preprint arXiv:2411.15098}, 2024.

\bibitem[Team(2025)]{hunyuanvideo2025}
Tencent Hunyuan Foundation~Model Team.
\newblock Hunyuanvideo 1.5 technical report, 2025.
\newblock URL \url{https://arxiv.org/abs/2511.18870}.

\bibitem[Wan et~al.(2025)Wan, Wang, Ai, Wen, Mao, Xie, Chen, Yu, Zhao, Yang,
  Zeng, Wang, Zhang, Zhou, Wang, Chen, Zhu, Zhao, Yan, Huang, Feng, Zhang, Li,
  Wu, Chu, Feng, Zhang, Sun, Fang, Wang, Gui, Weng, Shen, Lin, Wang, Wang,
  Zhou, Wang, Shen, Yu, Shi, Huang, Xu, Kou, Lv, Li, Liu, Wang, Zhang, Huang,
  Li, Wu, Liu, Pan, Zheng, Hong, Shi, Feng, Jiang, Han, Wu, and Liu]{wan2025}
Team Wan, Ang Wang, Baole Ai, Bin Wen, Chaojie Mao, Chen-Wei Xie, Di~Chen,
  Feiwu Yu, Haiming Zhao, Jianxiao Yang, Jianyuan Zeng, Jiayu Wang, Jingfeng
  Zhang, Jingren Zhou, Jinkai Wang, Jixuan Chen, Kai Zhu, Kang Zhao, Keyu Yan,
  Lianghua Huang, Mengyang Feng, Ningyi Zhang, Pandeng Li, Pingyu Wu, Ruihang
  Chu, Ruili Feng, Shiwei Zhang, Siyang Sun, Tao Fang, Tianxing Wang, Tianyi
  Gui, Tingyu Weng, Tong Shen, Wei Lin, Wei Wang, Wei Wang, Wenmeng Zhou, Wente
  Wang, Wenting Shen, Wenyuan Yu, Xianzhong Shi, Xiaoming Huang, Xin Xu, Yan
  Kou, Yangyu Lv, Yifei Li, Yijing Liu, Yiming Wang, Yingya Zhang, Yitong
  Huang, Yong Li, You Wu, Yu~Liu, Yulin Pan, Yun Zheng, Yuntao Hong, Yupeng
  Shi, Yutong Feng, Zeyinzi Jiang, Zhen Han, Zhi-Fan Wu, and Ziyu Liu.
\newblock Wan: Open and advanced large-scale video generative models.
\newblock \emph{arXiv preprint arXiv:2503.20314}, 2025.

\bibitem[Wang et~al.(2025)Wang, Yang, Zhao, Zhang, Liu, Zhou, and
  Xie]{wang2025gptimageedit15mmillionscalegptgeneratedimage}
Yuhan Wang, Siwei Yang, Bingchen Zhao, Letian Zhang, Qing Liu, Yuyin Zhou, and
  Cihang Xie.
\newblock Gpt-image-edit-1.5m: A million-scale, gpt-generated image dataset,
  2025.
\newblock URL \url{https://arxiv.org/abs/2507.21033}.

\bibitem[Wang et~al.(2004)Wang, Bovik, Sheikh, and Simoncelli]{wang2004image}
Zhou Wang, Alan~C Bovik, Hamid~R Sheikh, and Eero~P Simoncelli.
\newblock Image quality assessment: from error visibility to structural
  similarity.
\newblock \emph{IEEE transactions on image processing}, 13\penalty0
  (4):\penalty0 600--612, 2004.

\bibitem[Wiedemer et~al.(2025)Wiedemer, Li, Vicol, Gu, Matarese, Swersky, Kim,
  Jaini, and Geirhos]{wiedemer2025video}
Thadd{\"a}us Wiedemer, Yuxuan Li, Paul Vicol, Shixiang~Shane Gu, Nick Matarese,
  Kevin Swersky, Been Kim, Priyank Jaini, and Robert Geirhos.
\newblock Video models are zero-shot learners and reasoners.
\newblock \emph{arXiv preprint arXiv:2509.20328}, 2025.

\bibitem[Wu et~al.(2025{\natexlab{a}})Wu, Zheng, Yan, Xiao, Luo, Wang, Li,
  Jiang, Liu, Zhou, et~al.]{wu2025omnigen2}
Chenyuan Wu, Pengfei Zheng, Ruiran Yan, Shitao Xiao, Xin Luo, Yueze Wang, Wanli
  Li, Xiyan Jiang, Yexin Liu, Junjie Zhou, et~al.
\newblock Omnigen2: Exploration to advanced multimodal generation.
\newblock \emph{arXiv preprint arXiv:2506.18871}, 2025{\natexlab{a}}.

\bibitem[Wu et~al.(2025{\natexlab{b}})Wu, Ren, Shen, Cao, He, Lu, Gao, Xie,
  Lan, Alvarez, Gao, Fidler, Wang, and Ling]{wu2025chronoedit}
Jay~Zhangjie Wu, Xuanchi Ren, Tianchang Shen, Tianshi Cao, Kai He, Yifan Lu,
  Ruiyuan Gao, Enze Xie, Shiyi Lan, Jose~M. Alvarez, Jun Gao, Sanja Fidler,
  Zian Wang, and Huan Ling.
\newblock Chronoedit: Towards temporal reasoning for image editing and world
  simulation.
\newblock \emph{arXiv preprint arXiv:2510.04290}, 2025{\natexlab{b}}.

\bibitem[Wu et~al.(2025{\natexlab{c}})Wu, Huang, Wu, Cheng, Ding, and
  He]{Wu_2025_ICCV}
Shaojin Wu, Mengqi Huang, Wenxu Wu, Yufeng Cheng, Fei Ding, and Qian He.
\newblock Less-to-more generalization: Unlocking more controllability by
  in-context generation.
\newblock In \emph{Proceedings of the IEEE/CVF International Conference on
  Computer Vision (ICCV)}, pages 18682--18692, October 2025{\natexlab{c}}.

\bibitem[Xiao et~al.(2025)Xiao, Wang, Zhou, Yuan, Xing, Yan, Li, Wang, Huang,
  and Liu]{xiao2025omnigen}
Shitao Xiao, Yueze Wang, Junjie Zhou, Huaying Yuan, Xingrun Xing, Ruiran Yan,
  Chaofan Li, Shuting Wang, Tiejun Huang, and Zheng Liu.
\newblock Omnigen: Unified image generation.
\newblock In \emph{Proceedings of the IEEE/CVF Conference on Computer Vision
  and Pattern Recognition}, pages 13294--13304, 2025.

\bibitem[Xu et~al.(2024)Xu, Guo, Wang, Huang, Essa, and Shi]{xu2024prompt}
Xingqian Xu, Jiayi Guo, Zhangyang Wang, Gao Huang, Irfan Essa, and Humphrey
  Shi.
\newblock Prompt-free diffusion: Taking" text" out of text-to-image diffusion
  models.
\newblock In \emph{Proceedings of the IEEE/CVF Conference on Computer Vision
  and Pattern Recognition}, pages 8682--8692, 2024.

\bibitem[Xu et~al.(2026)Xu, Yan, Cao, Cheng, Hang, He, Yin, Zhang, Zhang, Li,
  Wang, Lu, Lee, and Tang]{xu2026tagmoetaskawaregatingunified}
Yu~Xu, Hongbin Yan, Juan Cao, Yiji Cheng, Tiankai Hang, Runze He, Zijin Yin,
  Shiyi Zhang, Yuxin Zhang, Jintao Li, Chunyu Wang, Qinglin Lu, Tong-Yee Lee,
  and Fan Tang.
\newblock Tag-moe: Task-aware gating for unified generative mixture-of-experts,
  2026.
\newblock URL \url{https://arxiv.org/abs/2601.08881}.

\bibitem[Yang et~al.(2024)Yang, Teng, Zheng, Ding, Huang, Xu, Yang, Hong,
  Zhang, Feng, et~al.]{yang2024cogvideox}
Zhuoyi Yang, Jiayan Teng, Wendi Zheng, Ming Ding, Shiyu Huang, Jiazheng Xu,
  Yuanming Yang, Wenyi Hong, Xiaohan Zhang, Guanyu Feng, et~al.
\newblock Cogvideox: Text-to-video diffusion models with an expert transformer.
\newblock \emph{arXiv preprint arXiv:2408.06072}, 2024.

\bibitem[Ye et~al.(2023)Ye, Zhang, Liu, Han, and Yang]{ye2023ip}
Hu~Ye, Jun Zhang, Sibo Liu, Xiao Han, and Wei Yang.
\newblock Ip-adapter: Text compatible image prompt adapter for text-to-image
  diffusion models.
\newblock \emph{arXiv preprint arXiv:2308.06721}, 2023.

\bibitem[Yu et~al.(2024)Yu, Chow, Yue, Pan, Wu, Wan, Li, Tang, Zhang, and
  Zhuang]{yu2024anyedit}
Qifan Yu, Wei Chow, Zhongqi Yue, Kaihang Pan, Yang Wu, Xiaoyang Wan, Juncheng
  Li, Siliang Tang, Hanwang Zhang, and Yueting Zhuang.
\newblock Anyedit: Mastering unified high-quality image editing for any idea.
\newblock \emph{arXiv preprint arXiv:2411.15738}, 2024.

\bibitem[Zhang et~al.(2023{\natexlab{a}})Zhang, Mo, Chen, Sun, and
  Su]{zhang2023magicbrush}
Kai Zhang, Lingbo Mo, Wenhu Chen, Huan Sun, and Yu~Su.
\newblock Magicbrush: A manually annotated dataset for instruction-guided image
  editing.
\newblock \emph{Advances in Neural Information Processing Systems},
  36:\penalty0 31428--31449, 2023{\natexlab{a}}.

\bibitem[Zhang et~al.(2023{\natexlab{b}})Zhang, Rao, and
  Agrawala]{zhang2023adding}
Lvmin Zhang, Anyi Rao, and Maneesh Agrawala.
\newblock Adding conditional control to text-to-image diffusion models.
\newblock In \emph{Proceedings of the IEEE/CVF international conference on
  computer vision}, pages 3836--3847, 2023{\natexlab{b}}.

\bibitem[Zhang et~al.(2024)Zhang, Song, Liu, Wang, Yu, Tang, Li, Tang, Hu, Pan,
  et~al.]{zhang2024ssr}
Yuxuan Zhang, Yiren Song, Jiaming Liu, Rui Wang, Jinpeng Yu, Hao Tang, Huaxia
  Li, Xu~Tang, Yao Hu, Han Pan, et~al.
\newblock Ssr-encoder: Encoding selective subject representation for
  subject-driven generation.
\newblock In \emph{Proceedings of the IEEE/CVF Conference on Computer Vision
  and Pattern Recognition}, pages 8069--8078, 2024.

\bibitem[Zhang et~al.(2025{\natexlab{a}})Zhang, Yuan, Song, Wang, and
  Liu]{zhang2025easycontrol}
Yuxuan Zhang, Yirui Yuan, Yiren Song, Haofan Wang, and Jiaming Liu.
\newblock Easycontrol: Adding efficient and flexible control for diffusion
  transformer.
\newblock \emph{arXiv preprint arXiv:2503.07027}, 2025{\natexlab{a}}.

\bibitem[Zhang et~al.(2025{\natexlab{b}})Zhang, Xie, Lu, Yang, and
  Yang]{zhang2025icedit}
Zechuan Zhang, Ji~Xie, Yu~Lu, Zongxin Yang, and Yi~Yang.
\newblock In-context edit: Enabling instructional image editing with in-context
  generation in large-scale diffusion transformers.
\newblock In \emph{Advances in Neural Information Processing Systems
  (NeurIPS)}, 2025{\natexlab{b}}.
\newblock arXiv:2504.20690.

\bibitem[Zhao et~al.(2026)Zhao, Xiong, Min, Hou, Zhang, and
  Jiang]{zhao2026condfoodgen}
Mengyao Zhao, Hao Xiong, Weiqing Min, Sujuan Hou, Mengmeng Zhang, and Shuqiang
  Jiang.
\newblock Condfoodgen: A conditional two-stream network for controllable food
  image generation.
\newblock \emph{IEEE Transactions on Image Processing}, 2026.

\end{thebibliography}
\endgroup
\end{document}